\documentclass[sigconf,nonacm]{acmart}
\AtBeginDocument{%
  }
\setcopyright{none}
\acmDOI{}
\acmISBN{}
\usepackage{pgffor}  
\usepackage{xcolor}
\usepackage{booktabs}
\usepackage{amsthm}
\usepackage{multirow}
\definecolor{michelleblue}{RGB}{0, 102, 204}
\definecolor{elenared}{RGB}{200, 0, 0}
\definecolor{jifanlightgreen}{RGB}{0, 180, 100}
\newcommand{\parent}[1]{\text{parent}(#1)}

\newcommand{\michelle}[1]{#1}
\newcommand{\elena}[1]{#1}
\newcommand{\jifan}[1]{#1}

\newcommand{\std}[1]{\scriptsize{$\pm$#1}}
\makeatletter
\newcommand{\myxspace}{%
  \@ifnextchar.{\relax}{%
  \@ifnextchar,{\relax}{%
  \@ifnextchar;{\relax}{%
  \@ifnextchar:{\relax}{%
  \@ifnextchar!{\relax}{%
  \@ifnextchar?{\relax}{%
  \@ifnextchar){\relax}{%
  \@ifnextchar]{\relax}{%
\@ifnextchar/{\relax}{\ }}}}}}}}}}
\makeatother

\newcommand{\iid}{\text{i.i.d.}\myxspace}
\newcommand{\name}{\textsc{GraCE-VAE}}
\newcommand{\bs}[1]{\boldsymbol{#1}}

\newtheorem{theorem}{Theorem}
\newtheorem{lemma}[theorem]{Lemma}
\newtheorem{proposition}[theorem]{Proposition}

\theoremstyle{definition}
\newtheorem{definition}[theorem]{Definition}
\newtheorem{assumption}[theorem]{Assumption}

\theoremstyle{remark}
\newtheorem{remark}[theorem]{Remark}

\begin{document}

\title{Causal Representation Learning from Network Data}

\author{Jifan Zhang}
\authornote{Both authors contributed equally to this research.}
\affiliation{%
  \department{Department of Statistics and Data Science,}
  \institution{Northwestern University,}
  \city{Evanston}
  \state{IL}
  \country{USA}
}
\email{jifanzhang2026@u.northwestern.edu}
\author{Michelle M.~Li}
\authornotemark[1]
\affiliation{%
      \department{Department of Biomedical Informatics,}
      \institution{Harvard University,}
  \city{Boston}
  \state{MA}
  \country{USA}
}
\email{michelleli@g.harvard.edu}
\author{Elena Zheleva}
\affiliation{%
  \department{Department of Computer Science,}
  \institution{University of Illinois Chicago,}
  \city{Chicago}
  \state{IL}
  \country{USA}}
\email{ezheleva@uic.edu}

\renewcommand{\shortauthors}{Zhang et al.}

\begin{abstract}
Causal disentanglement from soft interventions is identifiable under the assumptions of linear interventional faithfulness and availability
of both observational and interventional data. Prior work has focused on unstructured observations without leveraging known relational context among measured entities. In many scientific applications, however, the measured variables come with an observed interaction network that provides structured context, such as protein-protein interactions and pathway-gene membership. We propose GraCE-VAE, a graph-aware causal discrepancy variational autoencoder that treats pathway-level information as an auxiliary view of the latent causal programs. 
The graph neural network encoder conditions on this auxiliary pathway view and the biological graph to improve amortized inference, while the causal decoder remains a latent SCM with soft interventions. Assuming samples are i.i.d. within each intervention regime, we show that GraCE-VAE inherits the identifiability guarantees of causal discrepancy VAEs and identifies the latent causal graph and intervention targets up to the standard equivalence class.  Experiments on three CRISPR perturbation datasets demonstrate that leveraging structured biological context improves prediction of interventional outcomes, including unseen perturbation combinations.


\end{abstract}



\maketitle
\section{Introduction}
\label{sec:intro}

Discovering the causal factors underlying data and how they influence each other is a key challenge in machine learning. Early disentangled representation learning methods have aimed to learn latent features that correspond to distinct generative factors, typically assuming that these factors are statistically independent~\cite{higgins2018towards,kumar2017variational,higgins2017beta}. However, this assumption rarely holds in real-world scenarios, where generative factors often exhibit intricate dependencies. This realization has led to causal disentanglement approaches, which relax the independence assumption by modeling latent factors as nodes in a causal graph that can influence each other, i.e.~the structural causal model (SCM). Recent works have begun learning such causal representations from data, for instance, by integrating SCM constraints into variational autoencoders (VAEs)~\cite{yang2021causalvae,zhang2023identifiability}. These methods show that with appropriate data (including interventions) and assumptions, one can recover the latent causal model up to equivalence and predict intervention outcomes. Notably,~\citet{zhang2023identifiability} prove that even when the true causal variables are unobserved, a combination of observational and interventional data is sufficient to identify the latent causal directed acyclic graph (DAG) up to permutations of equivalent factors and to accurately extrapolate to novel interventions. Such progress provides a foundation for causal representation learning in complex settings.

While causal disentanglement has advanced for data represented as flat feature vectors, many real-world scientific datasets exhibit rich relational structure among the observed entities. Examples include social networks, where users are connected by friendships and hidden traits may influence their behaviors, and biological networks, where genes or proteins may interact via unobserved regulatory factors. \jifan{In these settings, the observed variables are associated with rich relational context, including links among measured entities (e.g., friendship ties, genetic interactions) and higher-level structure (e.g., communities, biological pathways). 
While such structured information may be informative of latent causal factors, existing causal representation learning methods largely ignore it.} Conversely, graph-based inference methods often assume that the causal graph is fully or partially known~\cite{feng2023concept,sanchez2021vaca}, or consider only observable interventions on known nodes~\cite{khemakhem2021causal}. This leaves a gap for learning latent causal structure and intervention effects while exploiting an observed context network among measured entities.

In this paper, we introduce a novel approach, Graph-Aware Causal Effects Variational AutoEncoder (\name), that combines graph neural networks with deep generative modeling to discover latent causal factors from network data enriched with interventions. At a high level, \name{} extends the VAE framework to incorporate an external network data's structure among the entities. The encoder of \name{} is a graph neural network (GNN) that processes the heterogeneous network data of observed nodes ($\mathbf{X}$-entities), their \jifan{auxiliary-view} nodes ($\mathbf{H}$-entities), and various relationships between them. This GNN-based encoder passes messages along the observed edges, capturing structural context (e.g.,~connectivity patterns and neighbor attributes) to produce enriched embeddings for the $\mathbf{X}$ nodes. These structure-aware embeddings are then fed into a variational inference network to approximate the posterior over latent variables. By doing so, the encoder infuses information from the network into the inferred latent representation, biasing the model towards explanations consistent with the relational patterns. The decoder of \name{} instantiates an SCM over the latent variables. In particular, the decoder learns a DAG that captures the hypothesized causal structure among the latent factors. 
Interventions are modeled through a two-stage process. First, an intervention encoder maps each intervention indicator to a soft selection over latent variables and a proposed mechanism modification. Then, the decoder applies this modification by altering the causal mechanism of the selected latent variable, while leaving others unchanged. \name{} simultaneously learns: (1)~the latent causal graph $G$, (2)~the correspondence of interventions to latent targets via the intervention encoder's assignments, and (3)~a generative model capable of simulating interventional outcomes on the observed network. Importantly, our identifiability statement is made under the standard assumption that samples are i.i.d. within each interventional regime; the network is used as structured side information rather than as a source of dependence across samples.
The work has \textbf{3 key contributions}:
\begin{itemize}
    \item We propose a framework for latent causal discovery that leverages structured context. \name, a variational graph autoencoder for causal disentanglement, jointly learns a latent causal graph and intervention effects that are informed by prior knowledge (via heterogeneous network of measured entities and group nodes). 
    Unlike methods that assume a given causal graph, fully observed variables, or independent observations, \name{} discovers the latent causal structure, guided by network connectivity and intervention data.
    
    \item We clarify the theoretical status of GraCE-VAE: since the generative model class matches prior causal discrepancy VAEs, and the graph-aware encoder only changes amortized inference, \name{} inherits the identifiability guarantees~\cite{zhang2023identifiability} under the same assumptions. We explicitly state the equivalence class and the i.i.d.-within-regime requirement.
    
    \item  Through experiments on three genetic perturbation datasets, we show that GraCE-VAE learns latent DAGs that support accurate interventional outcome prediction and provide biologically plausible hypotheses, especially for unseen perturbation combinations. Ablations show that both the graph-based encoder and the causal decoder are crucial to its success; removing either component degrades both structure learning and predictive accuracy. We also discuss how to interpret the learned latent DAGs and the limitations when interventions are missing or only observational data are available.
\end{itemize}

\name{} offers a new approach to uncovering latent causal factors in complex network systems, pushing the state-of-the-art in causal discovery and graph representation learning. It addresses a critical gap by unifying relational information and intervention-based causal inference, leading to more reliable causal analysis.

\section{Related Works}

\subsection{Causal disentanglement learning}

The concept of disentangled representation is formally defined by~\citet{higgins2018towards}. A representation is considered disentangled if it can be decomposed into independent features such that altering a single factor of the input data affects only one corresponding feature. Most disentangled representation learning methods are based on generative models, particularly VAEs~\cite{kumar2017variational,higgins2017beta,zhu2021commutative}. VAEs approximate the unknown true posterior 
$p(z|x)$ using a variational posterior $q(z|x)$. To enhance disentanglement, researchers have introduced various regularization techniques to the original VAE loss function. For instance, $\beta$-VAE~\cite{higgins2017beta} incorporates a penalty coefficient into the Evidence Lower Bound loss to enforce a stronger independence constraint on the posterior distribution 
$q(z|x)$.

Some studies have disagreed with the independence assumption and proposed causal disentanglement learning, assuming that the generating factors
of the observable data are causally influenced by the group
of confounding factors~\cite{suter2019robustly}.
SCM describes causal relationships
among generating factors~\cite{pearl2000causality}.
Numerous studies have investigated methods for identifying latent causal graphs by maximizing log-likelihood scores~\cite{chickering2002optimal,cussens2020gobnilp,solus2021consistency,raskutti2018learning}. Extensive research on causal disentanglement learning have focused on the discovery of causality from observational data~\cite{yang2021causalvae,kong2023identification,buchholz2024learning,varici2023score}. \citet{yang2021causalvae} propose CausalVAE, implementing the causal disentanglement process introduced by~\citet{suter2019robustly}. \citet{varici2023score} examine the identifiability of nonparametric latent SCMs under linear mixing, with the assumption of exactly one intervention per latent node. \citet{buchholz2024learning} investigate the identifiability of linear latent SCMs under nonparametric mixing, considering both hard and soft interventions within a framework of linear SCMs with additive Gaussian noise. 
\citet{zhang2023identifiability} study identifiability for soft interventions and offers a learning algorithm based on VAE without any structural restriction on the map from latent to observed variables. \name{} builds upon CMVAE~\cite{zhang2023identifiability} to incorporate structured context via a graph neural network. \jifan{Our theoretical contribution is therefore a scope clarification rather than a new identifiability theorem for a different decoder family: because \name{} keeps the CMVAE latent SCM decoder and modifies amortized inference, it inherits the same identifiability conditions while allowing the encoder to condition on an observed context graph.} 


\subsection{Variational autoencoders and graph neural networks}
GNNs excel in encoding the topology in network data, while VAEs provide a principled framework for learning latent representations through probabilistic inference. Their combination yields variational graph autoencoders (VGAEs) for network data~\cite{kipf2016variational}.

GNNs and VGAEs have since been used for causal discovery~\cite{khemakhem2021causal,sanchez2021vaca,tao2024deepite}. \citet{khemakhem2021causal} propose CAREFL, an autoregressive normalizing flow for causal discovery and inference on bivariate networks. However, their evaluation only considers interventions on root nodes, which simplifies to mere conditioning on those variables. \citet{tao2024deepite} design a VGAE for intervention target estimation that can concurrently learn across diverse causal graphs and sets of intervention targets in a self-supervised mode. \citet{sanchez2021vaca} propose VACA, a class of variational graph autoencoders for causal inference, utilizing GNNs to encode the causal graph information. \citet{feng2023concept} propose CCVGAE to obtain optimal disentangled
latent representations and conduct link prediction for the causal graph given partial causal graph structures. However, their settings~\cite{feng2023concept,sanchez2021vaca} assume that a causal graph over the variables is given. In contrast to all the settings mentioned above, \name{} learns a latent causal DAG and associates intervention labels to latent targets while using an observed context network as side information in the encoder. 

\subsection{Perturbation outcome prediction}
In genomics, related work focuses on predicting post-perturbation gene expression rather than recovering a latent causal graph. For example, GEARS \cite{roohani2024predicting} combines deep learning with gene-gene knowledge graphs to predict outcomes of single- and multi-gene perturbations, and scGen\cite{lotfollahi2019scgen} learns conditional generative models for perturbation response. These approaches typically treat perturbation targets as observed and do not aim to identify a latent causal DAG among unobserved factors. We view them as complementary: \name{} targets causal representation learning and counterfactual simulation via a latent SCM, while also being competitive on intervention-outcome prediction. \jifan{SENA~\cite{de2025interpretable}, a pathway-space causal representation learning method that incorporates biological prior knowledge, is closer to our setting than purely predictive perturbation-response models because it also aims to learn interpretable causal representations in pathway space. It differs from \name{} in how biological structure is used and in the combination of graph-aware inference with a latent SCM decoder.}  

\section{Problem Setup}
\subsection{Observed variables and context graphs}
{
We observe a heterogeneous context network with two entity types: $\mathbf{X}\in\mathbb R^d=(X_1,X_2,\cdots,X_d)$ denotes the primary observed view where each sample $x_n\in \mathbb R^d$ is a full observation vector over $d$ entities; and \jifan{auxiliary} entities $\mathbf{H}=(H_1,H_2,\cdots,H_m)$. Multiple relational edges connect entities within and across types, yielding adjacency structures $\mathbf A=(A_{XX},A_{XH},A_{HH})$. We refer to the graph ($\mathbf{X},\mathbf{H},\mathbf{A}$) as the context graph.
In our genomics experiments, \michelle{$\mathbf{X}$} are genes with expression values varying across samples, \michelle{$\mathbf{H}$} are pathways, and the edges encode protein-protein interactions, pathway-gene memberships, and pathway-pathway relations.}

\subsection{Data and sampling assumption}

{We study the setting in which both observational and interventional data are available. For each regime \michelle{$k \in \{0, 1, ..., K\}$ where $k=0$ denotes observational data and $k\geq1$ denotes intervention $I_k$, } 
we observe dataset $D^{(k)}$ consisting of $N_k$ samples of features on the $X$ entities, denoted $\{x_n^{(k)}\}_{n=1..N_k}$. 
\jifan{In the case where only $x_n^{(k)}$ is measured, \(h_n^{(k)}\) is not directly observed and the encoder uses the fixed pathway graph and random Gaussian for pathway-node embeddings as auxiliary pathway information.}

The standard i.i.d. assumption is applied at the sample level: samples within each regime are i.i.d. draws from $P_{X^{(k)}}$. This matches common perturbation screens where each cell is an independent sample.
} To ensure identifiability under the most challenging conditions, we assume that each latent variable is subject to at least one intervention. This assumption aligns with the identifiability requirement established for linear structural causal models, i.e.,~one intervention per node for recovery when one does not assume prior knowledge of the target of interventions  \( I_1, \dots, I_K \)~\citep{squires2023linear}.

\subsection{Latent SCM and interventions}
We posit that latent variables $\mathbf{U}=(U_1,U_2,\cdots,U_p)$ drive the generation of \michelle{$\mathbf{X}$} through an unknown mixing function $f$ such that $\mathbf{X}=f(\mathbf{U})$. Denote \( P_{X} \) as the induced distribution over \( \mathbf{X}\).

Denote $P_{U}$ as the joint distribution for $\mathbf{U}$, which can be factorized according to an unknown DAG \( G \) with $p$ nodes representing \( \{U_1, U_2\dots, U_p\} \). $G$ and $P_{U}$ are faithful to each other (i.e.,~every conditional independence that holds in $P_U$ corresponds exactly to a $d$-separation in 
$G$, and vice versa). Mathematically, 
\(
\mathbf{U} \sim P_{U} = \prod_{i=1}^{p} P(U_i \mid U_{\text{parent}_G(i)})\),
where \( \text{parent}_G(i) = \{j \in [p] : U_j \to U_i\} \) represents the index set of the parents of \( U_i \) in \( G \). 

We focus on single-node soft interventions applied to latent variables. Specifically, under an intervention \( I \) targeting a node \(U_i \), the original joint distribution \( P_U \) is altered to an interventional distribution \( P_U^{I} \). Similarly, denote the induced interventional distribution over \( \mathbf{X} \) as \( P_X^{I} \). This intervention modifies only the conditional distribution of the target node, replacing \( P(U_i \mid U_{\text{parent}_G(i)}) \) with a new conditional distribution \( P^{I}(U_i \mid U_{\text{parent}_G(i)}) \), while preserving the conditional distributions of all other (non-target) nodes. Formally, the factorization of the interventional distribution \( P_U^{I} \) is given by
\(
P_U^{I} = \prod_{i=1}^{p} P^{I}(U_i \mid U_{\text{parent}_G(i)})
\),
where \( P^{I}(U_j \mid U_{\text{parent}_G(j)}) = P(U_j \mid U_{\text{parent}_G(j)}) \) for all non-target nodes \(U_j\neq U_i  \). 

\subsection{Learning goals and scope}
{
Given an observational dataset $D^{(0)}$ and $K$ interventional datasets ${D^{(k)}}_{k=1..K}$, our objectives are to (i) learn a latent causal DAG $G$, (ii) associate each intervention label $I_k$ with its latent target, and (iii) predict outcomes of novel interventions, including new combinations. The latent variables \michelle{$\mathbf{U}$}, their dimensionality $p$, the DAG $G$, and the intervention targets are all treated as unobserved.}

\jifan{For identifiability, prior work requires that each latent variable is affected by at least one intervention and that the interventions satisfy linear interventional faithfulness. We adopt the same assumptions when stating our identifiability result. Even if some latent variables are never intervened upon, \name{} can still be trained and can remain useful for prediction, but the full latent DAG may not be identifiable (Section~\ref{limit}). }

\section{Model: \name}

\begin{figure*}[htbp]
    \centering
    \includegraphics[width=0.9\textwidth]{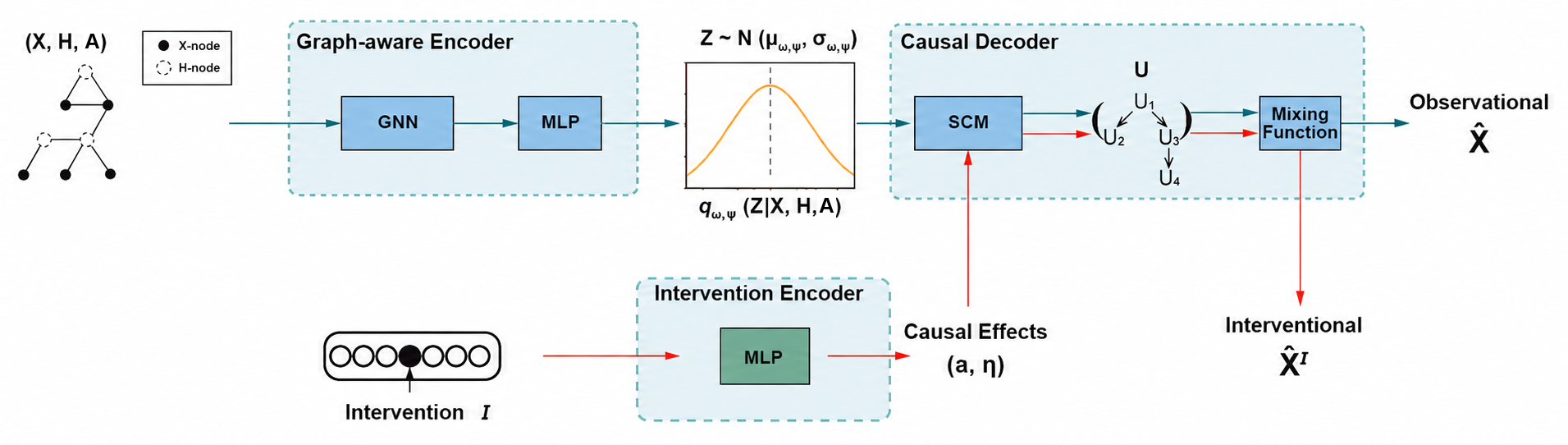}
    \caption{Overview of \name's architecture. \name{} consists of a graph-aware encoder, intervention encoder, and a causal decoder. Given entities $\mathbf{X}, \mathbf{H}$, context graph $\mathbf{A}$, and intervention $I$, \name{} learns a causal directed acyclic graph of the latent variables $\mathbf{U}$ to generate observational $\hat{\mathbf{X}}$ and interventional $\hat{\mathbf{X}^I}$ data.}
    \Description{\jifan{Architecture diagram of GraCE-VAE. The left side shows gene and pathway entities connected by a biological context graph and passed through a graph-aware encoder. The middle shows an intervention encoder that maps an intervention label to a soft latent target and effect. The right side shows an SCM decoder that learns a latent DAG over programs and generates observational and interventional gene-expression outputs.}}
    \label{fig:model}
\end{figure*}

\jifan{We propose \name\footnote{Code is available at \href{https://github.com/michellemli/GraCE-VAE}{github.com/michellemli/GraCE-VAE}}, a model that couples a GNN-based encoder with an SCM-based causal decoder for end-to-end discovery of latent causal graphs from heterogeneous network data and for prediction of intervention effects on the observed system. The key design is an encoder--decoder split: the biological network structure guides amortized inference, while the latent SCM decoder and the intervention encoder remain responsible for the learned causal generator.} Here, we describe \name's VAE framework for causal learning, \name's graph-aware encoder, \name's causal decoder, and the learning objective (Figure~\ref{fig:model}). 

\subsection{VAE framework for causal learning}  \label{sec:avb}

We adopt the VAE framework as the backbone for \name. Widely adopted in prior research on causal disentanglement~\cite{zhang2023identifiability,lippe2022citris,brehmer2022weakly}, the VAE enables a theoretically grounded and robust optimization process, with the advantages in scalability and flexibility in the choice of components. Our VAE has a latent prior $p(Z)$ and a decoder $p_{\theta}(X|Z)$, \jifan{whose likelihood reconstructs the gene expression view. 
The graph-aware inference network is allowed to condition on the expression view, the pathway-level auxiliary view, and the biological connections $q_{\omega,\psi}(Z\mid X,H,A).$ The training objective remains an \(X\)-marginal ELBO:}
\begin{align}
\log p_{\theta}(X) &
\ge
\mathbb E_{q_{\omega,\psi}(Z\mid X,H,A)}
\left[\log p_\theta(X\mid Z)\right]
\notag\\&-
\mathrm{KL}\!\left(
q_{\omega,\psi}(Z\mid X,H,A)\,\Vert\,p(Z)
\right).
\end{align}

For causal representation learning, we reinterpret the latent vector as a set of causal variables $\mathbf{U}=(U_1,\cdots,U_p)$ linked by an unknown DAG. Each conditional distribution is reparametrized as $$\mathbb{P}(U_i|U_{\parent{i}})=c_i(U_{\parent i},Z_i)$$ where $Z_i$ is an independent exogenous noise variable, and $c_i$ is the causal mechanism that generates $U_i$ from its parent(s) and $Z_i$. Under this view, the standard VAE likelihood $p_{\theta}(X|Z)$ factors into a deep SCM, $\mathbf{Z} \xrightarrow{c} \mathbf{U} \xrightarrow{f_\theta} \mathbf{X}$. The pathway-level view \michelle{\(\mathbf{H}\)} is used by the encoder to infer \michelle{\(\mathbf{U}\)}, but it is not an input to the \(X\)-decoder. With this reparameterization, VAE jointly learns the causal graph over $\mathbf{U}$, the causal mechanisms $c_i$, and the mixing function $f_{\theta}$.

\subsection{Graph-aware encoder}\label{sec:encoder}

To exploit the relational context defined by \michelle{\(\mathbf{A}\)}, we use a GNN as the encoder. For each sample, the nodes associated with the primary view \michelle{\(\mathbf{X}\)} are initialized with the corresponding observed attributes. For each $H$ node without observable features, we initialize the state with an \iid draw from a standard normal $N(0,I)$. These auxiliary features are used only by the encoder, together with the graph structure, to support graph-aware amortized inference.

\name{} is flexible with respect to the choice of GNN layer. Here, we instantiate a GraphSAGE layer~\cite{hamilton2017inductive} for concreteness (refer to Section~\ref{sec:gnn} for other types of GNNs). For a node \(v\in V\), GraphSAGE updates its representation by

\[
\widehat{r}_v \;=\;
\sigma\!\Biggl(
    W_{\text{self}}\,r_v
    \;+\;
    W_{\text{nbr}}\,
    \frac{1}{\lvert\mathcal N(v)\rvert}
    \sum_{u\in\mathcal N(v)} r_u
\Biggr),
\qquad v \in V,
\] where $\mathcal N(v)$ is the neighbor set of \(v\) in the context graph, $W_{\rm nbr}$ are learnable weights, and $\sigma(\cdot)$ is a non-linear activation. The GNN produces structure-aware embeddings for the nodes associated with \michelle{\(\mathbf{X}\)}. These embeddings are then passed to a variational inference head that parameterizes the posterior over the exogenous noise variables: $ q_{\omega,\psi}(Z\mid X,H,A)
    =
    \mathrm{MLP}_{\psi}
    \big(
    \mathrm{GNN}_{\omega}(X,H,A)
    \big),$ where \(\omega\) denotes the GNN parameters and \(\psi\) denotes the parameters of the variational head.

\jifan{The graph-aware encoder uses the context graph \((X,H,A)\) as side information for amortized inference. 
By passing messages over relations among the primary and auxiliary nodes, it produces structure-aware embeddings that parameterize the posterior over the exogenous noise variables \(Z\). 
Importantly, this context graph is not the latent causal graph itself and does not enter the \(X\)-decoder; it only improves the inference of latent programs while preserving the CMVAE generative family used for identifiability. 
The same encoder interface can also incorporate different sample representations, such as embeddings from single-cell foundation models, by concatenating or otherwise combining them with expression-derived node features before message passing.}

\subsection{Causal decoder and latent DAG}\label{sec:decoder}
Following~\citet{zhang2023identifiability}, the generative decoder $p_{\theta}(X|Z)$ factorizes into consecutive blocks of an SCM and a mixing function that mirror an implicit causal process. Given an observational sample and an intervention $I$, \name{} generates a counterfactual sample $\hat{x}^{I}$ as well as the reconstructed observational sample $\hat{x}$. These counterfactuals are compared against the true interventional data in the loss, enabling \name{} to learn both the latent DAG and the causal effects in an end-to-end manner.\\

\noindent\textbf{Structural Causal Model (SCM).} 
The exogenous noise vector $\mathbf{Z}\sim N(\bs \mu_{\omega,\psi}(X,H,A),\bs \sigma_{\omega,\psi}( X,H,A))$, where  $\bs\mu_{\psi,\omega}(\mathbf X),\bs\sigma_{\psi,\omega}(\mathbf X)$ are the mean and standard deviation vectors produced by $q_{\omega,\psi}$. Each coordinate of the noise vector $Z_i$, together with the values of its parents, deterministically produces a causal variable $U_i=c_i(U_{\parent i},Z_i),i\in[p]$.  The $c_i$ functions for soft interventions often take flexible forms, such as neural networks or shift functions that depend on $U_{\parent i}$, but introduce an altered mechanism rather than a hard fix~\cite{massidda2023causal}. However, for simplicity, we consider the MLP function.
The set $\parent i$ is defined by an \emph{upper-triangular} adjacency matrix $M_{\theta}$ that represents the directed edges in DAG $G$; sparsity is promoted through the penalty term $-\lambda||M_{\theta}||_1$.

To model an intervention $I$, an intervention encoder $(a,\eta)=T_{\phi}(I)$ outputs $a\in \Delta^{p-1}$, a soft one-hot vector where the largest entry selects the target
latent node $v$ and $\eta$ that specifies the new mechanism for the
intervention. We then inform an interventional set of mechanisms: \[
c^{I}_{i}\!\bigl(U_{\!\operatorname{pa}(i)},Z_i\bigr)=
\begin{cases}
c_{i}\!\bigl(U_{\!\parent i},Z_i\bigr), & i\neq v,\\[6pt]
g_{i}^{\eta}\!\bigl(U_{\!\parent i},Z_i\bigr), & i = v,
\end{cases}
\qquad i \in [p].
\notag
\] $g_i^{\eta}$ can be an arbitrary learnable function; here we instantiate 
$g_i^{\eta}$ as an additive shift for computational simplicity: $$g_i^{\eta}=c_i(U_{\parent i},Z_i)+\eta,\eta\in R.$$
\jifan{This intentionally simple intervention encoder learns a soft latent target assignment and an intervention effect without hard-coding a perturbation graph into the decoder. We also evaluate more complex intervention encoders that capture additional priors.} \\

\noindent\textbf{Mixing Function.} A neural map $f_{\theta}: \mathbb{R}^p\to \mathbb{R}^d$ converts the causal vector to the observed space $\mathbf{X}=f_{\theta}(\mathbf{U})$. The same mixing function $f_{\theta}$ is shared by observational and interventional regimes. 

\subsection{Objective function}\label{sec:obj}

\name{} is trained on both observational data $P(X)$ and $K$ interventional data $P^{I_k}(X),k\in[K]$. The optimization target therefore couples an {ELBO} term and a {distribution--alignment} term. For observational samples, the ELBO is:
\begin{align*}
L_{\omega,\theta,\psi}^{\text{ELBO}}:=&\mathbb E_{x,h\sim P_{\mathrm{obs}}(X,H)}
[
-
\mathbb E_{q_{\omega,\psi}(z\mid x,h,A)}\left[
\log p_\theta(x\mid z)
\right]\\&+
\beta\,
\mathrm{KL}
\left(
q_{\omega,\psi}(z\mid x,h,A)\,\Vert\,p(z)
\right)].
\end{align*}
\jifan{When pathway-level observations \(h\) are unavailable, \(h\) is initialized by standard Gaussian, yielding the implemented encoder.}

The distribution-alignment term forces the decoder to reproduce each interventional distribution once the corresponding mechanism has been replaced by the intervention encoder. Formally, let $\hat{X}^{I_k}=f_{\theta}(U^{I_k})$ be the model-generated counterfactuals: $$L_{\omega,\theta,\psi,\phi}^{\text{alignment}}:=\alpha\sum_{k=1}^K D_{\gamma}\bigl( P^{I_k}(X),P_{\omega,\theta,\psi,\phi}(\hat{X}^{I_k})\bigr)+\lambda\|M_{\theta}\|_1.$$
$D_{\gamma}$ is a discrepancy measurement between two distributions. 
Specifically, we penalize the maximum mean discrepancy (MMD) between the model-generated and empirical distributions for each intervention regime, and we add sparsity regularization $\|M_{\theta}\|$. 

We optimize a weighted sum of (i) the ELBO terms for observational and interventional samples, (ii) MMD discrepancy terms that align interventional distributions, and (iii) graph regularizers:
$$L_{\omega,\theta,\psi,\phi}=L^{ELBO}_{\omega,\theta,\psi}+L_{\omega,\theta,\psi,\phi}^{\text{alignment}},$$
where $\alpha,\beta,\gamma$ are hyperparameters. 

\subsection{Identifiability guarantee}\label{sec:identifia}

We analyze identifiability as prior causal VAE work with latent causal variables and soft interventions.
In particular, the latent causal structure is identifiable only up to an equivalence class:
two solutions are considered equivalent if they induce the same family of observational and
interventional distributions over \michelle{\(\mathbf{X}\)}.
Our framework builds on the identifiability theory of Zhang et al.~\cite{zhang2023identifiability}
for discrepancy-based causal VAEs (CMVAE). \jifan{Importantly, the identifiability argument is based on the \(X\)-marginal generative model. The encoder conditions on the context graph,
\(
    q_{\omega,\psi}(Z\mid X,H,A)
\),
but we assume that the decoder remains
\(
p_\theta(X\mid U,H,A)=p_\theta(X\mid U)
\).
Thus, \michelle{\(\mathbf{H}\)} and \michelle{\(\mathbf{A}\)} affect only amortized inference and do not change the latent SCM decoder or the soft-intervention model over \michelle{\(\mathbf{U}\)}.} Consequently, under the CMVAE assumptions on the \(X\)-marginal model and the additional auxiliary-view assumption, \name{} inherits the following identifiability guarantees.
\begin{proposition}[GraCE-VAE inherits \(X\)-marginal identifiability from CMVAE]
Assume that:
(i) for each intervention regime \(I_k\), the joint distribution admits the multi-view factorization
$$
    p^{I_k}(u,x,h\mid A)
    =
    p_U^{I_k}(u)\,p_\theta(x\mid u)\,p_\eta(h\mid u,A);
$$
(ii) the \(X\)-marginal model \(p_U^{I_k}(u)p_\theta(x\mid u)\) satisfies the latent SCM, decoder-rank, and soft-intervention assumptions of Zhang et al. ~\cite{zhang2023identifiability};
(iii) samples are i.i.d. within each regime;
(iv) each latent variable is intervened upon at least once and the interventions satisfy linear interventional faithfulness; and
(v) the variational family \(q_{\omega,\psi}(z\mid x,h,A)\) is sufficiently expressive so that, at the population optimum of the \(X\)-marginal objective, the ELBO is tight for the decoder parameters. Then, any global optimum of the population GraCE-VAE objective identifies the latent causal DAG over \michelle{\(\mathbf{U}\)} and the intervention targets up to the same equivalence class as in Zhang et al. \cite{zhang2023identifiability}.
\end{proposition}
\jifan{The additional auxiliary-view assumption \(p_\theta(X\mid U,H,A)=p_\theta(X\mid U)\) ensures that, for the \(X\)-marginal decoder, all information from \michelle{\(\mathbf{H}\)} relevant to generating \michelle{\(\mathbf{X}\)} is already summarized by latent state \michelle{\(\mathbf{U}\)}. \elena{For example, in \michelle{genomics}, this would hold if latent programs induce both the formation of pathways and gene expressions: \michelle{$\mathbf{U} \to \mathbf{H}, \mathbf{U} \to \mathbf{X}$}.
Then, the pathway information is used on the inference side to improve inference of \michelle{\(\mathbf{U}\)}, while the \(X\)-decoder and latent intervention mechanisms remain aligned with the CMVAE formulation.} If the additional auxiliary-view assumption fails, \name{} remains a valid predictive architecture, but we no longer claim inherited identifiability of the latent DAG or intervention targets.} In the appendix, a complete proof follows adapted from the corresponding identifiability results in
Zhang et al.~\cite{zhang2023identifiability}.


\section{Experimental Setup}
\subsection{Datasets}

We conduct experiments on three genetic perturbation datasets: \textsc{norman}, \textsc{replogle-small}, and \textsc{replogle-large}.

\textsc{norman} consists of an observational dataset $\mathcal{D}_0$ with 8,907 unperturbed cells and an interventional dataset with 99,590 perturbed cells. Cells are perturbed using CRISPR-based screens that target one (i.e.,~single perturbation) or two (i.e.,~double perturbation) of 105 genes. These interventions are organized into $K=217$ distinct interventional datasets, $\mathcal{D}_1,\mathcal{D}_2,\cdots,\mathcal{D}_K$, each with 50-2000 cells. A cell is represented by a 5000-dimensional vector $X$ that captures the expression of 5000 highly variable genes.

The two \textsc{replogle} datasets are derived from a large-scale CRISPR-based screen that targets 2,058 genes~\cite{replogle2022mapping}. \textsc{replogle-large} consists of an observational dataset $\mathcal{D}_0$ with 10,691 unperturbed cells and an interventional dataset with 299,694 perturbed cells resulting from single perturbations. Each cell is represented by a 8563-dimensional vector $X$ that captures the expression of 8,563 highly variable genes. At inference, we predict the effects of 414 perturbation targets (i.e.,~exclude perturbation targets with fewer than 200 cells). While \textsc{replogle-large} includes all perturbation targets during training, \textsc{replogle-small} contains only the 414 perturbation targets. As a result, its interventional dataset has 126,807 cells.



We define \textbf{structured biological knowledge} as pathway-gene and pathway-pathway associations~\cite{jassal2020reactome}, and physical protein-protein interactions~\cite{li2024contextual}. For \textsc{norman}, the structured context network consists of 2,694 pathway nodes and 5,000 protein nodes. It includes 22,786 edges connecting pathways to proteins, 2,713 edges between pathways, and 6,191 edges representing protein interactions.
For both \textsc{replogle} datasets, the structured context network consists of 2,694 pathway nodes and 8,563 protein nodes. It includes 75,845 edges connecting pathways to proteins, 2,713 edges between pathways, and 102,087 edges representing protein interactions. 

\begin{table*}  \small \centering \begin{tabular}{@{}llccccc@{}} \toprule Dataset & Metric & CMVAE & CMVAE-multihot & VGAE & SENA & \textbf{\name} \\ \midrule \multirow{6}{*}{\textsc{norman}} & Single R\textsuperscript{2} & 0.9018\std{0.0210} & 0.9338\std{0.0098} & \textbf{0.9493\std{0.0123}} & 0.8378\std{0.0181} & \underline{0.9452\std{0.0108}} \\ & Single RMSE & 0.4823\std{0.0031} & 0.4753\std{0.0057} & 0.4763\std{0.0045} & \textbf{0.4657\std{0.0238}} & \underline{0.4711\std{0.0066}} \\ & Single MMD & 0.1538\std{0.0188} & 0.1235\std{0.0106} & \textbf{0.1075\std{0.0146}} & 0.2513\std{0.0285} & \underline{0.1138\std{0.0126}} \\ & Double R\textsuperscript{2} & 0.7382\std{0.0209} & \underline{0.7844\std{0.0161}} & 0.7168\std{0.0640} & 0.6638\std{0.0992} & \textbf{0.7845\std{0.0241}} \\ & Double RMSE & 0.5228\std{0.0038} & \textbf{0.5175\std{0.0056}} & 0.5404\std{0.0095} & 0.6255\std{0.0788} & \underline{0.5204\std{0.0037}} \\ & Double MMD & 0.7990\std{0.0495} & \textbf{0.7071\std{0.0388}} & 0.8869\std{0.1916} & 1.1096\std{0.3261} & \underline{0.7090\std{0.0529}} \\ \bottomrule \end{tabular} \caption{Single- and double-intervention performance on \textsc{norman}. Best results are bolded and second-best results are underlined.} \label{tab:combined_results} \end{table*}

\begin{table*}
\small
\centering
\begin{tabular}{@{}llcccc@{}}
\toprule
Dataset & Metric & CMVAE & CMVAE-multihot & VGAE & \textbf{\name}\\\midrule
\multirow{3}{*}{\textsc{replogle-small}}
& Single R\textsuperscript{2} & 0.8318\std{0.0108} & \underline{0.8379\std{0.0139}} & 0.8347\std{0.0099} & \textbf{0.8388\std{0.0128}} \\
& Single RMSE                 & 0.6032\std{0.0135} & \underline{0.5849\std{0.0168}} & \textbf{0.5772\std{0.0051}} & 0.5869\std{0.0115} \\
& Single MMD                  & \textbf{0.5414\std{0.0090}} & \underline{0.5610\std{0.0223}} & 0.5970\std{0.0067} & 0.5836\std{0.0460} \\\midrule
\multirow{3}{*}{\textsc{replogle-large}}
& Single R\textsuperscript{2} & \underline{0.8346\std{0.0194}} & --- & 0.8018\std{0.0946} & \textbf{0.8493\std{0.0058}} \\
& Single RMSE                 & 0.5997\std{0.0165} & --- & \textbf{0.5875\std{0.0517}} & \underline{0.5948\std{0.0094}} \\
& Single MMD                  & \textbf{0.5251\std{0.0225}} & --- & 0.6750\std{0.0238} & \underline{0.5389\std{0.0113}} \\
\bottomrule
\end{tabular}
\caption{Single-intervention performance on \textsc{replogle-large} and \textsc{replogle-small}. Best results are bolded and second-best results are underlined.}
\label{tab:combined_results2}
\end{table*}
\subsection{Setup}
For \textsc{norman}, we randomly select 96 and 64 cells from each single intervention dataset ($\mathcal{D}_1,\cdots,\mathcal{D}_{105}$) containing more than 800 cells for the test and validation sets, respectively. All double intervention datasets are exclusively reserved for the test set. For \textsc{replogle-large} and \textsc{replogle-small}, we randomly split $10\%$ and $20\%$ for the validation and test sets. The remaining $70\%$ are used for training; only \textsc{replogle-large} includes the 1,644 perturbation targets with fewer than 200 samples in the train set.


For \textbf{baselines}, we evaluate against CMVAE, the state-of-the-art causal discrepancy VAE architecture~\cite{zhang2023identifiability}; CMVAE-multihot, a modification of CMVAE in which a multi-hot encoding of biological pathway information (where 1 indicates that any of the genes in a pathway is expressed in the cell, 0 otherwise) is used as additional input features; SENA~\cite{de2025interpretable}; and VGAE, an ablation of \name{} with a standard MLP decoder (instead of a causal decoder). In other words, CMVAE does not incorporate structured biological knowledge, unlike CMVAE-multihot (via multihot encoding) and VGAE (via GNN encoder). Due to computational and memory constraints, we exclude CMVAE-multihot from the evaluation on the \textsc{replogle-large} dataset and evaluate SENA only on the \textsc{norman} dataset. 

\jifan{We conduct \textbf{ablation} studies along three dimensions. We vary the GNN architecture (i.e.,~GCN~\cite{bhatti2023deep}, GAT~\cite{velivckovic2017graph}, GraphSAGE~\cite{hamilton2017inductive}) and the number of GNN layers $l \in {1,3}$ \michelle{(Table~\ref{tab:gnn_architecture})}. We vary the level of structured biological context by using different subsets of gene-gene, pathway-gene, and pathway-pathway edges \michelle{(Table~\ref{tab:level_of_structure})}. To assess whether the observed gains come from meaningful biological topology, we introduce a corrupted-graph ablation in which 50\% of the biological edges are replaced with random edges \michelle{(Table~\ref{tab:random_edge_ablation})}.} \michelle{We assess the contributions of the observation-side priors using embeddings from a single-cell foundation model. We also evaluate alternative intervention encoders for capturing intervention-side priors: Tri-value initializes the features of the target gene, candidate target gene, and non-target gene nodes as 1, 0, or -1, respectively, before encoding the graph; Tri-value + 1-hop additionally restricts the graph to the perturbed genes and their one-hop neighborhoods; DropEdge drops edges of the perturbed genes (Table~\ref{tab:ablation_prior}).} 



%

We use standard \textbf{metrics} to compare the ground truth and generated perturbation effects after single or double intervention: Maximum Mean Discrepancy (MMD), $R^2$, and Root Mean Squared Error~(RMSE). Following prior work~\cite{zhang2023identifiability}, we compute these metrics on the differentially expressed genes (DEGs), which exhibit significant changes in expression under interventions. \jifan{This DEG-focused evaluation avoids diluting the perturbation signal with weakly or non-responsive genes, and better reflects the recovery of biologically meaningful responses. This choice is also consistent with recent discussions that standard correlation-based and distributional metrics in single-cell perturbation prediction can be sensitive to scale, sparsity, and dimensionality, making DEG-level evaluation a more targeted and interpretable criterion~\cite{heidari2026evaluating}.}

\jifan{For reproducibility, all reported means and standard deviations are computed over 10 random seeds with matched data splits across methods. Hyperparameters, such as $\alpha$, $\beta$, $\lambda$, learning rate, are selected via the validation set; details on hyperparameter sweeps, compute setup, and code availability are provided in the appendix.}

\section{Results}

We evaluate \name's performance to answer the following research questions. \textbf{R1:}~What is the contribution of structured context for predicting the effects of single interventions? \textbf{R2:}~What is the contribution of structured context for predicting the effects of unseen intervention combinations? \textbf{R3:}~How do different graph neural networks capture structured context? \textbf{R4:}~How does the amount of prior knowledge (via structured context) affect performance? 
\michelle{\textbf{R5:}~How does biologically-meaningful graph topology affect performance? \textbf{R6:}~To what extent do observation- or intervention-side priors improve generalization?}

\subsection{\texorpdfstring{\underline{R1}: Structured context for single interventions}{R1: Structured context for single interventions}}

We evaluate the contributions of structured context for predicting the effects of single-interventions on \textsc{norman} and \textsc{replogle}.

Across the three datasets, \name{} achieves strong $R^2$ performance, obtaining the best $R^2$  on \textsc{replogle-small}, \textsc{replogle-large}, and \textsc{norman} double interventions, while remaining competitive on \textsc{norman} single interventions (Tables~\ref{tab:combined_results}-\ref{tab:combined_results2}). Notably, \name{} achieves an average R\textsuperscript{2} of $0.9452$ compared to CMVAE's $0.9018$ on \textsc{norman}. CMVAE-multihot and VGAE, which also consider structured context, outperform CMVAE with average R\textsuperscript{2} of $0.9338$ and $0.9493$, respectively. \jifan{While \name{} does not win at every single-intervention metric (e.g., in \textsc{norman}, VGAE has the strongest $R^2$ and MMD, while \name{} has the best RMSE), it has strong single-intervention performance and superior extrapolation to unseen double interventions.}

Compared with CMVAE-multihot, GraCE-VAE improves $R^2$ in most settings and achieves comparable or competitive performance on RMSE and MMD, while using a more compact graph-aware parameterization. This underscores the effectiveness of explicitly leveraging the topology of network data rather than relying solely on multihot-encoded features. VGAE, an ablation of \name, achieves comparable RMSE but worse R\textsuperscript{2} and MMD.

It is worth noting that models that consider structured context (i.e.,~\name, CMVAE-multihot, VGAE) outperform those that do not (i.e.,~CMVAE) in most metrics. \name's strong performance against all baseline models (particularly in R\textsuperscript{2}) demonstrates the advantage of combining structured context via a GNN architecture and causal modeling.

\subsection{\texorpdfstring{\underline{R2}: Structured context for double interventions}{R2: Structured context for double interventions}}

We assess the contributions of structured context for predicting double intervention effects in \textsc{norman}. Models are trained only on single interventions, and evaluated on unseen double interventions.

For unseen intervention combinations, the structured context learned by \name's GNN significantly enhances performance (Table~\ref{tab:combined_results}). \name{} achieves an average R\textsuperscript{2} of $0.7845$ and MMD of $0.7090$, surpassing CMVAE (average R\textsuperscript{2} of $0.7382$ and MMD of $0.7990$) and SENA (average R\textsuperscript{2} of $0.6638$ and MMD of $1.1096$).

\name{} performs comparably to CMVAE-multihot, emphasizing the benefit of incorporating structured context. Notably, this comparable performance is achieved with a more compact parameterization, as \name{} avoids parameter inflation introduced by large multihot encodings (Appendix Table~\ref{tab:parameter}). Even with fewer parameters, \name's encoding of structured knowledge yields comparable predictive performance on double interventions.

Compared to VGAE, the presence of a causal decoder in \name{} significantly improves its ability to generalize to novel intervention combinations, leading to higher R\textsuperscript{2} and lower RMSE and MMD. These results strongly indicate that the synergy between the GNN's encoding of structured context and the causal decoding in \name{} is critical for accurately and efficiently modeling and generalizing to complex, previously unseen interventions.

\begin{figure}[t]
\centering
\includegraphics[width=0.65\linewidth]{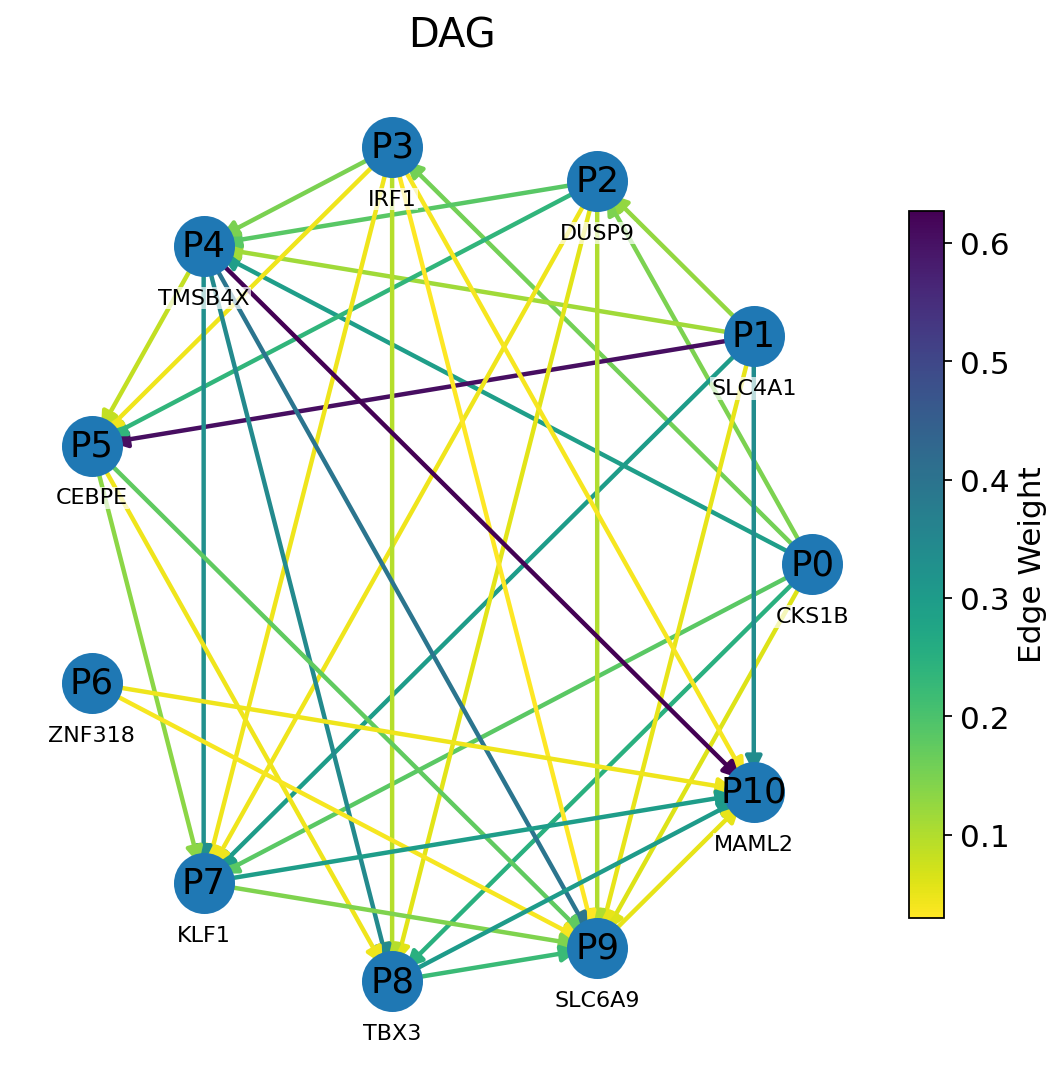}
\caption{DAG of the learned latent program by \name{} for \textsc{norman}. Each node is annotated with its most representative gene and edge weights correspond to the absolute values of the learned DAG coefficients.}
\Description{\jifan{Directed graph visualization of the learned \textsc{norman} latent programs. The figure contains 11 latent program nodes labeled by representative genes. Directed edges connect programs, and edge thickness or intensity represents the absolute magnitude of learned DAG coefficients. The graph is used to inspect plausible regulatory relationships among latent programs.}}
\label{fig:dag}
\end{figure}
\subsection{\texorpdfstring{\underline{R3}: Design of the graph neural network}{R3: Design of the graph neural network}}\label{sec:gnn}

Any GNN architecture can be flexibly integrated into \name. We evaluate different designs of GNN architectures on both single and double interventions in \textsc{norman}.

Recall that \name{} in Table~\ref{tab:combined_results} consists of a single GraphSAGE layer. Regardless of the GNN design (Table~\ref{tab:gnn_architecture}), \name{} with a 1- or 3-layer GCN, GAT, or GraphSAGE encoder significantly outperform CMVAE (Table~\ref{tab:combined_results}).

Comparing the different GNN architectures, \name{} with a GraphSAGE layer has the strongest performance on both single and double interventions. Meanwhile, GCN and GAT have comparable performance on most metrics. It seems that increasing the number of layers does not lead to performance improvement. This indicates that deeper architectures may not necessarily enhance the model's capacity to learn the underlying mechanisms.

\begin{table}[ht]
\small
\centering
\begin{tabular}{@{}lccc@{}}
\toprule
Metric & 1-layer GCN & 1-layer GAT & 3-layer GAT \\
\midrule
Single R\textsuperscript{2} & 0.9376\std{0.0106} & 0.9319\std{0.0107} & 0.9302\std{0.0133} \\
Single RMSE                 & 0.4744\std{0.0073} & 0.4725\std{0.0026} & 0.4762\std{0.0043} \\
Single MMD                  & 0.1242\std{0.0113} & 0.1274\std{0.0110} & 0.1306\std{0.0136} \\
Double R\textsuperscript{2} & 0.7593\std{0.0252} & 0.7594\std{0.0204} & 0.7609\std{0.0158} \\
Double RMSE                 & 0.5218\std{0.0031} & 0.5192\std{0.0016} & 0.5197\std{0.0044} \\
Double MMD                  & 0.7750\std{0.0619} & 0.7912\std{0.0531} & 0.7750\std{0.0340} \\
\bottomrule
\end{tabular}
\caption{Ablation of GNN in \name{} on \textsc{norman}.}
\label{tab:gnn_architecture}
\end{table}

\subsection{\texorpdfstring{\underline{R4}: Levels of structured context}{R4: Levels of structured context}}

While we have demonstrated strong performance gains due to incorporating structured context, we interrogate the contributions to predictive ability by leveraging different levels of structured context. Specifically, we train \name{} models that consider only gene-gene edges (GG); both gene-gene and pathway-gene edges (GG+PG); or gene-gene, pathway-gene, and pathway-pathway edges (GG+PG+PP). Note that pathway-pathway edges (PP) are limited to those between pathways that are connected to the genes in $\mathbf{X}$, whereas the pathway-pathway edges learned by \name{} in Table~\ref{tab:combined_results} are between all known pathways.


Introducing progressively richer edge information markedly boosts \name's performance on predicting the effects of single interventions (Tables~\ref{tab:level_of_structure}). Adding pathway--gene edges increases R\textsuperscript{2} and decreases RMSE on \textsc{norman} and \textsc{replogle-small}, and decreases MMD on \textsc{norman}. Also, the performance gains plateau (rather than reverse) when pathway--pathway edges are included.

The benefit of adding more structured context is less clear for \name's performance on predicting the effects of unseen intervention combinations (Table~\ref{tab:level_of_structure}). R\textsuperscript{2} is slightly decreased, accompanied by a minor increase in MMD, when additional edges are included. These results suggest a trade-off between contextual richness and overfitting risk when incorporating higher order pathway-level information.

We retain the complete network data in \name{} because its net effect across tasks is still favorable. The slight performance decrease in predicting the effects of double interventions remains within acceptable variance. Further, the full biological topology better positions the framework for future settings in which multi-step or higher-order perturbations become essential. Nevertheless, the inclusion of different levels of structured context in \name{} can be determined based on the domain or dataset.

\begin{table}[ht]
\small
\centering
\begin{tabular}{@{}llccc@{}}
\toprule
Dataset & Metric & GG & GG+PG & GG+PG+PP \\
\midrule
\multirow{6}{*}{\textsc{norman}}
& Single R\textsuperscript{2} & 0.9327\std{0.0205} & 0.9414\std{0.0117} & 0.9420\std{0.0119} \\
& Single RMSE & 0.4771\std{0.0044} & 0.4766\std{0.0034} & 0.4752\std{0.0075} \\
& Single MMD & 0.1270\std{0.0229} & 0.1186\std{0.0091} & 0.1233\std{0.0132} \\
& Double R\textsuperscript{2} & 0.7905\std{0.0191} & 0.7852\std{0.0175} & 0.7851\std{0.0177} \\
& Double RMSE & 0.5177\std{0.0053} & 0.5182\std{0.0035} & 0.5188\std{0.0050} \\
& Double MMD & 0.6990\std{0.0489} & 0.7073\std{0.0397} & 0.7094\std{0.0362} \\
\midrule
\multirow{3}{*}{\textsc{replogle}}
& Single R\textsuperscript{2} & 0.8326\std{0.0112} & 0.8401\std{0.0096} & 0.8414\std{0.0075} \\
&Single RMSE & 0.5927\std{0.0090} & 0.5898\std{0.0149} & 0.5900\std{0.0112} \\
& Single MMD & 0.5624\std{0.0091} & 0.5691\std{0.0237} & 0.5683\std{0.0192} \\
\bottomrule
\end{tabular}
\caption{Ablation of the structured context captured by \name{} on \textsc{norman} and \textsc{replogle-small}.} 
\label{tab:level_of_structure}
\end{table}

\subsection{\texorpdfstring{\underline{R5}: Robustness to corrupted graph structure}{R5: Robustness to corrupted graph structure}}

\jifan{To assess whether the improvement comes from biologically meaningful topology rather than merely adding edges or parameters, we replace 50\% of the biological edges in \textsc{norman} with random edges while keeping the graph size comparable.}
\michelle{We find that corrupting edges consistently degrades performance relative to the full biological graph across single- and double-intervention metrics (Table~\ref{tab:random_edge_ablation}).} 
\jifan{The drop is especially clear for single-intervention $R^2$ and MMD, supporting the interpretation that meaningful biological structure helps graph-aware inference.}
\begin{table}[h]
\small
\centering
\begin{tabular}{@{}lcc@{}}
\toprule
Metric & \textbf{\name} & \jifan{Random 50\% Edge} \\
\midrule
Single R\textsuperscript{2} & \textbf{0.9452\std{0.0108}} & \jifan{0.9271\std{0.0125}} \\
Single RMSE                 & \textbf{0.4711\std{0.0066}} & \jifan{0.4767\std{0.0050}} \\
Single MMD                  & \textbf{0.1138\std{0.0126}} & \jifan{0.1329\std{0.0128}} \\
Double R\textsuperscript{2} & \textbf{0.7845\std{0.0241}} & \jifan{0.7791\std{0.0210}} \\
Double RMSE                 & \textbf{0.5204\std{0.0037}} & \jifan{0.5218\std{0.0038}} \\
Double MMD                  & \textbf{0.7090\std{0.0529}} & \jifan{0.7183\std{0.0476}} \\
\bottomrule
\end{tabular}
\caption{\jifan{Ablation of structure context edges on \textsc{norman}.}} 
\label{tab:random_edge_ablation}
\end{table}

\subsection{\texorpdfstring{\underline{R6}: Priors and modularity}{R6: Priors and modularity}}

\begin{table*}[!ht]
\small
\centering
\begin{tabular}{@{}lccccc@{}}
\toprule
Metric 
& \jifan{Tri-value} 
& \jifan{Tri-value + 1-hop} 
& \jifan{DropEdge} 
& \textbf{\name} 
& \jifan{\name{} + scFM} \\
\midrule
Single R\textsuperscript{2} 
& \jifan{\underline{0.9448\std{0.0166}}} 
& \jifan{0.9392\std{0.0144}} 
& \jifan{0.8539\std{0.0183}} 
& \textbf{0.9452\std{0.0108}} 
& \jifan{0.9324\std{0.0105}} \\

Single RMSE                 
& \jifan{\underline{0.4652\std{0.0089}}} 
& \jifan{\textbf{0.4639\std{0.0065}}} 
& \jifan{0.4795\std{0.0175}} 
& 0.4711\std{0.0066} 
& \jifan{0.4775\std{0.0046}} \\

Single MMD                  
& \jifan{\underline{0.1170\std{0.0241}}} 
& \jifan{0.1236\std{0.0134}} 
& \jifan{0.2089\std{0.0180}} 
& \textbf{0.1138\std{0.0126}} 
& \jifan{0.1250\std{0.0068}} \\

Double R\textsuperscript{2} 
& \jifan{0.7181\std{0.0541}} 
& \jifan{0.7069\std{0.0350}} 
& \jifan{0.6512\std{0.0421}} 
& \textbf{0.7845\std{0.0241}} 
& \jifan{\underline{0.7794\std{0.0209}}} \\

Double RMSE                 
& \jifan{0.5297\std{0.0159}} 
& \jifan{0.5273\std{0.0066}} 
& \jifan{0.5254\std{0.0125}} 
& \underline{0.5204\std{0.0037}} 
& \jifan{\textbf{0.5190\std{0.0037}}} \\

Double MMD                  
& \jifan{0.9015\std{0.1412}} 
& \jifan{0.9395\std{0.0909}} 
& \jifan{1.0660\std{0.1604}} 
& \textbf{0.7090\std{0.0529}} 
& \jifan{\underline{0.7130\std{0.0409}}} \\
\bottomrule
\end{tabular}
\caption{\michelle{Ablation of observation- and intervention-side priors on \textsc{norman}.}}
\label{tab:ablation_prior}
\end{table*}

\michelle{To evaluate how \name's observation- and intervention-side priors affect performance, we implement alternative approaches.}

\michelle{\name's graph-aware encoder is modular with respect to additional observation-side priors. To test \name's ability to capture sufficiently predictive observation-side priors from graph and cell features, we concatenate} 
\jifan{embeddings from TranscriptFormer, a single-cell foundation model (scFM), with the original gene expression vectors on \textsc{norman}.} \michelle{We find that the resulting model (i.e.,~\name{} + scFM) only improves on double RMSE; the original \name{} remains the strongest model (Table~\ref{tab:ablation_prior}).} 
\jifan{We also assess alternative intervention-side priors.} 
\michelle{While Tri-value, Tri-value + 1-hop, and DropEdge can be competitive for some single-intervention metrics, the default \name{} intervention encoder is the strongest on unseen double interventions (Table~\ref{tab:ablation_prior})}. 
\michelle{These experiments reinforce that \name's improvements do not come from simply injecting more context or graph structure.} 


\subsection{Interpretability and robustness}

Because the true latent causal graph among unobserved factors is not available in real genomics datasets, we avoid claiming exact graph recovery. Instead, we view the learned DAG as a hypothesis space that supports counterfactual simulation and can be inspected for biological plausibility. Edges in the learned DAG can then be evaluated by checking whether connected latent programs show known regulatory or pathway-level relationships. We also note that identifiability results require that each latent variable is targeted by at least one intervention; when this assumption is violated, \name{} can still be trained for prediction, but parts of the latent DAG may become underdetermined. Finally, with only observational data, \name{} reduces to a graph-aware VAE and can learn representations, but causal directions are generally unidentifiable without additional assumptions.

\jifan{To suggest a biological interpretation of the learned DAG, we inspect the 11 latent programs for \textsc{norman}. Program P0, represented by genes such as CKS1B, TBX2, PTPN12, ZBTB25, BPGM, and SET, and pathways such as R-HSA-1640170 and R-HSA-1280215, corresponds to a cell-cycle regulatory module. P1, with genes such as SLC4A1, NCL, DLX2, COL2A1, SNAI1, SET, ZC3HAV1, and COL1A1 and pathways such as R-HSA-1474244 and R-HSA-9758919, is consistent with extracellular-matrix remodeling and mesenchymal identity. P2, the largest module, seems to act as a central cell-fate decision hub. P3--P10 respectively align with interferon signaling, proliferative signaling in hematopoietic progenitors, myeloid lineage commitment, zinc-finger transcriptional regulation, erythroid differentiation, maintenance of undifferentiated or stem-cell states, glycine transport and chondrogenic identity, and stress-responsive apoptotic integration. These annotations suggest that the learned DAG provides a biologically coherent hematopoietic differentiation hypothesis space, although they should be viewed as qualitative support rather than definitive causal validation.}

\section{Limitations}\label{limit}
Our framework, \name, relies on several assumptions that may limit its applicability in practice.
First, our formulation and identifiability discussion assume access to interventional data and sufficient coverage of interventions across latent causal variables.
When interventions are weak, entangled, or violate the single-target soft-intervention assumption, the learned latent causal structure may become ambiguous.
Second, for computational simplicity, we model interventions using a restricted functional form (i.e.,~additive shifts in the causal mechanisms). Such formulations may not fully capture complex biological perturbation effects.
\jifan{Third, our identifiability statement relies on the auxiliary-view assumption: pathway-level information \michelle{$\mathbf{H}$} may help infer the latent causal programs, but once \michelle{\(\mathbf{U}\)} is given, it does not provide additional generative information for the \michelle{\(\mathbf{X}\)}. 
If \michelle{$\mathbf{H}$} contain information about \michelle{\(\mathbf{X}\)} beyond what is captured by \michelle{\(\mathbf{U}\)}, then the inherited identifiability guarantee no longer applies.}
Finally, our empirical evaluation focuses on genetic perturbation datasets and curated biological networks. While these benchmark datasets are widely used, generalization to other domains or network types requires further validation.

\section{Conclusion}

\jifan{We introduce \name, a framework for learning latent causal models from perturbation data with observed structured biological context. \name{} uses the biological graph as auxiliary information for graph-aware amortized inference, while an SCM decoder learns the latent DAG, intervention-target assignments, and interventional outcomes. This encoder--decoder split allows the model to exploit biological context without hard-coding the curated pathway graph as the causal generator, and preserves the inherited CMVAE identifiability guarantee under the i.i.d.-within-regime, intervention-coverage, linear interventional-faithfulness and auxiliary-view assumptions. Empirically, \name{} is the strongest model in the most scientifically meaningful setting: predicting unseen double interventions after training only on single interventions. Future directions include developing stronger quantitative validation of latent causal programs against external perturbation, lineage, and pathway evidence; extending the theory to graph-conditioned decoders, sample-specific graphs, and genuinely dependent networked samples; and designing principled observation-side and intervention-side priors that incorporate biological knowledge without sacrificing combinatorial generalization.}



\begin{acks}
This research was supported in part by the U.S. National Science Foundation under Grants No. 2047899 and 2217023. Any opinions, findings, and conclusions or recommendations expressed in this material are those of the author(s) and do not necessarily reflect the views of the NSF or the U.S. Government.
\end{acks}

\bibliographystyle{ACM-Reference-Format}
\bibliography{reference}
\newpage
\appendix
\section{Ethical Use of Data and GenAI Disclosure}
This work uses publicly available genetic perturbation datasets and biological knowledge bases released for research use. We did not collect new data, recruit human participants, perform human-subject interventions, or access private or identifiable personal information; thus, this study does not constitute human-subjects research under IRB policies, and no IRB approval was required. We used ChatGPT by OpenAI only for manuscript preparation, including proofreading and alternative phrasing; all scientific content, experiments, results, and conclusions were produced and verified by the authors, and no GenAI tool was used to generate or modify data, labels, or experimental results. Model-predicted causal relations or intervention effects should not be treated as clinical or biomedical evidence without experimental validation and domain oversight.

\section{Implementation details}

\begin{table*}[h]
\small
\centering
\begin{tabular}{@{}lllll@{}}
\toprule
\textbf{Dataset}                  & \name & CMVAE & CMVAE-multihot& VGAE  \\ \toprule
 \sc{norman} & 1,364,035       & 1,364,029
       & 1,630,531&1,380,120       \\ 
       \sc{replogle-small} &  2,639,093   & 2,639,087  & 2,983,919
       &  2,574,220     \\       \sc{replogle-large} &  2,498,987      & 
        2,498,981& 2,843,699&  2,500,804        \\ \bottomrule
\end{tabular}
\caption{Number of parameters in \name{} and baselines.}
\label{tab:parameter}
\end{table*}
Models are implemented in PyTorch and PyTorch Geometric. All models are trained on the same data splits across 10 seeds $s \in [1, 10]$ on the validation dataset. Each model is run on one GPU. We independently perform hyperparameter sweeps for each baseline, GNN architecture ablation, and \name. The hyperparameters are selected from: $\alpha_{\max} \in \{8, 10, 12, 15\}$, $\beta_{\max} \in \{2, 3, 4\}$, $\lambda\in \{0.01,0.001, 0.0001\}$, and \texttt{Temp} $\in \{4, 5, 6\}$. For \textsc{replogle-large}, the selection of latent dimensions $p \in \{100, 200, 300\}$. The best hyperparameter configurations are selected based on R\textsuperscript{2}, MMD, and RMSE.  Table~\ref{tab:hyperparameters_split} and Table~\ref{tab:hyperparameter} summarize the detailed information for each model. We will explain in detail the meaning of each coefficient in the following paragraphs.

We use a composite loss function: (1)~a reconstruction loss based on log likelihood, (2)~a KL divergence regularization, (3)~an MMD term to align generated samples with observed data distributions, and (4)~a regularization term to encourage sparsity of the latent DAG. The MMD loss is computed using a mixture of Gaussian kernels with dyadically spaced bandwidths, following \citet{gretton2012kernel}. The base kernel bandwidth is set to $\sigma = 1000$  and the kernel widths are chosen geometrically as ${\sigma, 2\sigma, \ldots, 2^{\texttt{kernel\_num}-1}\sigma}$, with $\texttt{kernel\_num} = 10$. This multi-scale kernel approach helps stabilize training and ensures sensitivity to discrepancies across a range of scales. The coefficients for the MMD loss ($\alpha$) and KL loss ($\beta$) are scheduled during training as follows: $\alpha$ is set to 0 for the first 5 epochs, then linearly increased to $\alpha_{\max}$ over the next half of training, and held constant thereafter. $\beta$ is set to 0 for the first 10 epochs, then increased linearly to $\beta_{\max}$ by the midpoint of training and kept fixed for the remaining epochs. The coefficient ($\lambda$) for sparsity regularization term is fixed throughout training.

For VAE parametrization, the encoder consists of GNN layer(s) followed by two fully connected layers with 128 hidden units each, while the decoder consists of a DAG layer followed by two fully connected layers with 128 hidden units. We use leaky ReLU activations for the encoder and decoder. Intervention encoding is handled by a neural network $T_\phi$ that maps each intervention one-hot indicator to a soft one-hot assignment over the latent space. We employ an annealing schedule for the softmax temperature: $t = 1$ for the first half of epochs, and then $t$ is linearly increased to \texttt{Temp}.

\begin{table*}[h]
\small
\centering
\begin{tabular}{@{}llllll@{}}
\toprule
Dataset&Hyperparameter        & \name &CMVAE & CMVAE-multihot&VGAE\\ \midrule
\multirow{8}{*}{\textsc{norman}}
&Latent dimension & 105&105&105&105\\ &Number of pathways & 2694& 2694& 2694& 2694\\
&\(\alpha_{\max}\)                & 8   &12&8& 8        \\
&\(\beta_{\max}\)                 & 2      &4&3&2      \\
&$\lambda$            & 0.0001  &0.001&0.0001& 0.0001     \\
&\texttt{Temp}               & 4        &6&4&  4    \\ \midrule
\multirow{8}{*}{\textsc{replogle-small}}
&Latent dimension & 414 & 414& 414& 414\\ &Number of pathways & 2694 &2694 &2694 &2694\\
&\(\alpha_{\max}\)                & 12       & 12& 15&  12    \\
&\(\beta_{\max}\)                 & 3      &2 & 2&  3      \\
&$\lambda$            & 0.0001  &0.0001 &0.0001 &    0.0001  \\
&\texttt{Temp}              & 5      &6 & 6&    5     \\ \midrule
\multirow{8}{*}{\textsc{replogle-large}}
&Latent dimension & 300 & 300& ---& 300\\ &Number of pathways & 2694 &2694 &--- &2694\\
&\(\alpha_{\max}\)                & 12       & 12& ---&  12    \\
&\(\beta_{\max}\)                 & 3      &2 & ---&  3    \\
&$\lambda$            & 0.0001  &0.0001 &--- &    0.0001      \\
&\texttt{Temp}              & 5      &6 & ---&    5     \\ \midrule
\end{tabular}
\caption{Best hyperparameter configurations for \name{} and baselines.}
\label{tab:hyperparameters_split}
\end{table*}

\begin{table*}[h]
\small
\centering
\begin{tabular}{@{}llll@{}}
\toprule
Hyperparameter        &1-layer GCN & 1-layer GAT& 3-layer GAT\\ \midrule

Latent dimension & 105&105&105\\ Number of pathways & 2694& 2694& 2694\\
\(\alpha_{\max}\)                & 8   &8& 8       \\
\(\beta_{\max}\)                 & 2      &4&3     \\
$\lambda$            & 0.001  &0.0001&0.0001     \\
\texttt{Temp}               &5        &6&5    \\\bottomrule
\end{tabular}
\caption{Best hyperparameter configurations for each GNN architecture ablation, evaluated on \textsc{norman}.}
\label{tab:hyperparameter}
\end{table*}

For all biological datasets, the choice of latent dimension $p$ is guided by the number of unique intervention targets and validated performance. For $\textsc{norman}$, we set $p=105$, matching the number of distinct intervention targets. For $\textsc{replogle-small}$, we set $p=414$, correspondingly. For $\textsc{replogle-large}$, we conduct a grid search and select  $p=300$. 
All models are trained using the Adam optimizer with learning rate $=0.001$. We use a batch size of $32$ and train each model for 100 epochs. Our code is publicly available at \href{https://github.com/michellemli/GraCE-VAE}{github.com/michellemli/GraCE-VAE}.

\section{Identifiability guarantee}
\label{sec:identifiability}
\paragraph{Scope and auxiliary-view reduction.}
Our identifiability statement concerns the \emph{
$X$-marginal} generative model.  Let $A$ denote the fixed observed context graph used by the encoder.  For each intervention regime $I_k$, we assume the auxiliary-view factorization
\begin{equation}
    p^{I_k}(u,x,h\mid A)
    =
    p_U^{I_k}(u)\,p_\theta(x\mid u)\,p_\eta(h\mid u,A).
    \label{eq:aux-view-factorization-app}
\end{equation}
Equivalently, once the latent causal state $U$ is given, the auxiliary view $H$ does not further change the $X$-decoder:
\begin{equation}
    p_\theta(x\mid u,h,A)=p_\theta(x\mid u),
    \qquad\text{or}\qquad
    X\perp\!\!\!\perp H\mid U,A.
    \label{eq:aux-view-ci-app}
\end{equation}
Thus $H$ may help the encoder infer $U$, but it is not an additional parent of $X$ in the reduced-form decoder.  Marginalizing over $H$ gives
\begin{equation}
    p^{I_k}_\theta(x)
    =
    \int p_\theta(x\mid u)p_U^{I_k}(u)\,du,
    \label{eq:x-marginal-app}
\end{equation}
which is the same $X$-marginal decoder family used in CMVAE.  When the decoder is written in the deterministic form used by the identifiability theory of \citet{zhang2023identifiability}, $X=f(U)$ and
\begin{equation}
    P_X^{I_k}=f_{\#}P_U^{I_k}.
    \label{eq:pushforward-app}
\end{equation}
Here $A$ is fixed within the dataset and is used by the encoder as graph structure; it is not an input to the reduced-form $X$-decoder.  The graph-aware encoder may condition on $H$ and $A$ through $q_{\omega,\psi}(u\mid x,h,A)$, but this is part of the inference model only and does not change either the $X$-decoder or the intervention mechanisms on $U$.  If the auxiliary-view assumption in~\eqref{eq:aux-view-factorization-app} fails, \name{} can still be trained and evaluated as a predictive architecture, but the inherited CMVAE identifiability guarantee below no longer applies.

All identifiability statements below are about the population $X$-marginal distributions
$P_X$ and $\{P_X^{I_k}\}_{k=1}^K$.  Finite samples are assumed i.i.d. within each intervention regime.

\subsection{Assumptions and equivalence class}

\begin{assumption}[Full-rank decoder and latent support]
\label{assump:app-fullrank}
The observable variables $\mathbf X\in\mathbb R^d$ are generated by a full-row-rank polynomial decoder
$f:\mathbb R^p\to\mathbb R^d$ via
\[
    \mathbf X=f(\mathbf U),
\]
and the interior of the support of $P_U$ is non-empty.
\end{assumption}

\begin{assumption}[Linear interventional faithfulness]
\label{assump:app-lif}
Denote by $\mathrm{ch}_G(i)$, $\mathrm{de}_G(i)$, and $\mathrm{an}_G(i)$ the children, descendants, and ancestors of $i$ in $G$.  A single-node intervention $I$ with target $i$ satisfies \emph{linear interventional faithfulness} if for every
$j\in\{i\}\cup \mathrm{ch}_G(i)$ such that $\mathrm{pa}_G(j)\cap \mathrm{de}_G(i)=\emptyset$, it holds that
\[
    \mathbb P\!\left(U_j+U_S c^\top\right)
    \neq
    \mathbb P^I\!\left(U_j+U_S c^\top\right)
    \quad\text{for all constants } c\in\mathbb R^{|S|},
\]
where $S=[p]\setminus(\{j\}\cup \mathrm{de}_G(i))$.
\end{assumption}

\begin{assumption}[Total separation]
\label{assump:app-ts}
For every edge $i\to j$ in $G$, there do not exist constants $c_j$ and $\{c_k\}_{k\in S}$ such that
\[
U_i \perp\!\!\!\perp
U_j+c_jU_i
\ \Big|\
\{U_\ell\}_{\ell\in\mathrm{pa}_G(j)\setminus(S\cup\{i\})},
\ \{U_k+c_kU_i\}_{k\in S},
\]
where $S=\mathrm{pa}_G(j)\cap \mathrm{de}_G(i)$.
\end{assumption}

\begin{definition}[CD-equivalence]
\label{def:app-cd}
Two triples
\[
\langle \mathbf U,G,I_1,\ldots,I_K\rangle,
\qquad
\langle \widehat{\mathbf U},\widehat G,\widehat I_1,\ldots,\widehat I_K\rangle
\]
are \emph{CD-equivalent} if there exist a permutation $\pi\in S_p$ and scalars $\lambda_i\neq 0$, $b_i\in\mathbb R$ such that
\[
    \widehat U_i=\lambda_iU_{\pi(i)}+b_i,
    \qquad
    \widehat G=G_\pi,
    \qquad
    \widehat I_k=(I_k)_\pi\quad(k=1,\ldots,K).
\]
\end{definition}

\subsection{Observational identifiability}

\begin{lemma}[Identifiability up to affine transformations]
\label{lem:app-affine}
Suppose Assumption~\ref{assump:app-fullrank} holds.  Then the latent dimension $p$ is identifiable from $P_X$.  Moreover, any alternative latent representation $\widehat{\mathbf U}$ that yields the same $P_X$ through a full-row-rank polynomial decoder satisfies
\[
    \widehat{\mathbf U}=\Lambda \mathbf U+b
\]
for some non-singular matrix $\Lambda\in\mathbb R^{p\times p}$ and vector $b\in\mathbb R^p$.
\end{lemma}

\begin{proof}
This is the observational part of the identifiability argument of \citet{zhang2023identifiability}.  Briefly, let $p^\ast$ be the smallest latent dimension for which there exist a latent distribution $P_{\widehat U}$ and a full-row-rank polynomial decoder $\widehat f:\mathbb R^{p^\ast}\to\mathbb R^d$ with $\widehat f(\widehat U)\overset{d}{=}X$.  Since both $f$ and $\widehat f$ are full-row-rank polynomial maps, their monomial expansions imply that the monomial vector of $\widehat U$ is an affine function of the monomial vector of $U$, and conversely.  Thus $\widehat U$ and $U$ are related by polynomial maps that are inverses on an open set.  Under the full-row-rank polynomial constraint, this inverse relation reduces to an affine reparameterization, so $\widehat U=\Lambda U+b$ with $\Lambda$ invertible.  Hence $p^\ast=p$.
\end{proof}

\subsection{Identifiability of the latent DAG and intervention targets}

\begin{theorem}[GraCE-VAE inherits $X$-marginal identifiability from CMVAE]
\label{thm:app-identifiability}
Assume the auxiliary-view factorization~\eqref{eq:aux-view-factorization-app}.  Assume also that the $X$-marginal model satisfies Assumptions~\ref{assump:app-fullrank}, \ref{assump:app-lif}, and~\ref{assump:app-ts}; that each latent variable is intervened upon at least once,
\[
    \bigcup_{k=1}^K \mathrm{TG}(I_k)=[p];
\]
and that the variational family $q_{\omega,\psi}(z\mid x,h,A)$ is sufficiently expressive so that, at the population optimum of the $X$-marginal objective, the ELBO is tight for the decoder parameters.  Then any global optimum of the population GraCE-VAE objective identifies the latent causal DAG over $\mathbf U$ and the intervention targets up to the same CD-equivalence class as in \citet{zhang2023identifiability}.
\end{theorem}

\begin{proof}
The auxiliary-view assumption is used only for the following reduction.  By~\eqref{eq:aux-view-factorization-app}, marginalizing over $H$ yields the same $X$-marginal observational and interventional distributions as a CMVAE model:
\[
    p_\theta^{I_k}(x)
    =
    \int p_\theta(x\mid u)p_U^{I_k}(u)\,du,
    \qquad k=0,1,\ldots,K.
\]
Equivalently, in the deterministic mixing-map notation of \citet{zhang2023identifiability},
\[
    X=f(U),
    \qquad
    U\sim P_U^{I_k},
    \qquad
    P_X^{I_k}=f_\# P_U^{I_k}.
\]
Therefore $H$ and the fixed graph $A$ do not alter the $X$-decoder family or the latent soft-intervention model.  The population distributions used for identification are exactly the $X$-marginal observational and interventional distributions considered in CMVAE.  The proof now follows the CMVAE/Zhang argument on this $X$-marginal family.

Let
$\langle \widehat{\mathbf U},\widehat G,\widehat I_1,\ldots,\widehat I_K\rangle$
be any alternative latent SCM and intervention assignment that induces the same $X$-marginal observational and interventional distributions.

\paragraph{Step 1: observational reduction to affine ambiguity.}
By Lemma~\ref{lem:app-affine}, $p$ is identifiable and
\begin{equation}
    \widehat{\mathbf U}=\Lambda \mathbf U+b
    \label{eq:app-affine-ambiguity}
\end{equation}
for some invertible matrix $\Lambda$ and vector $b$.

\paragraph{Step 2: intervention targets and transitive closure.}
Under Assumption~\ref{assump:app-lif} and the coverage condition $\cup_{k=1}^K\mathrm{TG}(I_k)=[p]$, Theorem~1 of \citet{zhang2023identifiability} implies that the intervention targets and the transitive closure $\mathrm{TS}(G)$ are identifiable up to a single permutation of the latent coordinates.  Concretely, there exists a topological order $\tau$ of $G$ such that, after reindexing by $\tau$, one obtains a sparsest topological representation
\begin{equation}
    \widehat U=U\widehat\Gamma+\widehat c,
    \label{eq:app-topological-repr}
\end{equation}
where $\widehat\Gamma$ is invertible and respects the descendant structure of $G$: for all $i<j$,
\begin{equation}
    \widehat\Gamma_{\tau(j),i}=0,
    \qquad
    \widehat\Gamma_{\tau(j),j}\neq 0,
    \qquad
    \tau(\ell)\notin \mathrm{de}_G(\tau(j))
    \Rightarrow
    \widehat\Gamma_{\tau(j),\ell}=0.
    \label{eq:app-descendant-sparsity}
\end{equation}
At this stage the recovered graph is the transitive closure $\mathrm{TS}(G_\tau)$, and the interventions are matched to their targets up to the same permutation.

\paragraph{Step 3: recovering direct edges.}
It remains to upgrade identification from the transitive closure to the direct DAG.  Fix the order $\tau$.  For any invertible upper-triangular matrix $R'$, define
\[
    \bar U:=\widehat U R'.
\]
Because $R'$ is upper triangular and $\widehat\Gamma$ satisfies~\eqref{eq:app-descendant-sparsity}, the representation $\bar U=U\bar\Gamma+\bar c$ retains the same descendant sparsity pattern.  Define a DAG $\bar G_{R'}$ on nodes $[p]$ by declaring, for $i<j$,
\begin{equation}
    i\to j\in \bar G_{R'}
    \quad\Longleftrightarrow\quad
    \bar U_i\not\!\perp\!\!\!\perp \bar U_j
    \ \Big|\
    \bar U_1,\ldots,\bar U_{i-1},\bar U_{i+1},\ldots,\bar U_{j-1}.
    \label{eq:app-graph-from-ci}
\end{equation}
Let $\bar R$ minimize the number of edges in $\bar G_{R'}$ over all invertible upper-triangular $R'$, and write $\bar G:=\bar G_{\bar R}$.

We show $\bar G=G_\tau$.  First, suppose an edge $\tau(i)\to\tau(j)$ in $G$ with $i<j$ were missing from $\bar G$.  Then~\eqref{eq:app-graph-from-ci} gives a conditional independence between $\bar U_i$ and $\bar U_j$ given the intervening coordinates.  Using the upper-triangular form, this conditional independence can be rewritten as
\begin{align}
U_{\tau(i)}\ \perp\!\!\!\perp\ U_{\tau(j)}+c_jU_{\tau(i)}
\ \Big| &\
U_{\tau(1)},\ldots,U_{\tau(i-1)},
U_{\tau(i+1)}\notag\\&+c_{i+1}U_{\tau(i)},\ldots,U_{\tau(j-1)}+c_{j-1}U_{\tau(i)}
\label{eq:app-derived-ci}
\end{align}
for some constants $c_{i+1},\ldots,c_j$.  Combining~\eqref{eq:app-derived-ci} with the local Markov property of $G$ and the intersection property of conditional independence yields
\begin{align}
U_{\tau(i)}\ \perp\!\!\!\perp\ U_{\tau(j)}+c_jU_{\tau(i)}
\ \Big|&\
\{U_{\tau(\ell)}\}_{\tau(\ell)\in \mathrm{pa}_G(\tau(j))\setminus(S\cup\{\tau(i)\})},
\{U_{\tau(\ell)}\notag\\&+c_\ell U_{\tau(i)}\}_{\tau(\ell)\in S},\notag
\end{align}
where $S=\mathrm{pa}_G(\tau(j))\cap\mathrm{de}_G(\tau(i))$.  This contradicts the total separation assumption for the edge $\tau(i)\to\tau(j)$.  Hence $G_\tau\subseteq \bar G$.

Conversely, because the representation is invertible and upper triangular in the order $\tau$, there exists an upper-triangular transform $R_0$ such that
\[
    \widehat U R_0=(U_{\tau(1)},\ldots,U_{\tau(p)})+c_0
\]
for some constant vector $c_0$.  For this transform, the local Markov property implies that if $\tau(i)$ is not a parent of $\tau(j)$, then
\[
U_{\tau(i)}\ \perp\!\!\!\perp\ U_{\tau(j)}
\ \Big|\
U_{\tau(1)},\ldots,U_{\tau(i-1)},U_{\tau(i+1)},\ldots,U_{\tau(j-1)},
\]
so $i\to j\notin \bar G_{R_0}$.  Hence $\bar G_{R_0}\subseteq G_\tau$.  Since $\bar R$ was chosen to minimize the number of edges, $|\bar G|\le |\bar G_{R_0}|\le |G_\tau|$.  Together with $G_\tau\subseteq\bar G$, this gives $\bar G=G_\tau$.

Thus the latent DAG and intervention targets are identified up to the same permutation found in Step~2, with the remaining ambiguity being coordinate-wise scaling and shifting.  This is exactly CD-equivalence.

Finally, the expressiveness and tightness assumption on $q_{\omega,\psi}(z\mid x,h,A)$ connects the population GraCE-VAE optimum to the same $X$-marginal decoder parameters and interventional distributions.  Since the encoder may condition on $H$ and $A$ without modifying the $X$-decoder, any global optimum of the population GraCE-VAE objective inherits the CMVAE identifiability conclusion.
\end{proof}

\paragraph{Implication for \name}
The theorem identifies the latent DAG over $U$ and the intervention targets from the $X$-marginal interventional family.  It does not claim identifiability of the auxiliary mechanism $p_\eta(h\mid u,A)$, causal relations among auxiliary nodes, or the fixed context graph itself.  The role of $H$ and $A$ is to improve amortized inference through the graph-aware encoder while preserving the CMVAE generative family used for the identifiability argument.

\section{Visualization of Generated Samples}

To visualize and compare the fidelity of the generated and ground-truth samples, we aggregate the control cells, ground-truth post-intervention cells, and model-generated samples into a single data matrix. Specifically, we concatenate the control expression matrix $X_{\text{ctrl}}$, the observed intervention data, and the generated outputs, and then construct a unified AnnData object. Each sample is annotated as 'NA' (control), 'Actual Cells' (real samples), or 'Generated Cells' (generated samples).

Dimensionality reduction is performed using PCA followed by UMAP, with the neighborhood graph constructed from the top 50 principal components. This two-step embedding projects the high-dimensional gene expression profiles into two dimensions, enabling direct visual comparison between the distributions of the real (or ground-truth) and generated samples. In each plot, samples are colored according to their label ('Actual Cells', 'Generated Cells', or 'NA'). We provide both an overall UMAP for all interventions as well as per-target UMAPs that highlight each single or double gene intervention individually. These visualizations enable qualitative assessment of how closely the generated samples match the true data distributions across various perturbation scenarios.

In our visual analyses, we focus on comparing \name{} and CMVAE, as these two models are representative of structured and unstructured approaches, respectively. By concentrating on these methods, we aim to highlight the impact of incorporating structured network knowledge while maintaining clarity in our figures.

\begin{figure*}[htbp]
  \centering
  \foreach \i in {1,...,4} {
    \begin{minipage}[b]{0.4\textwidth}
      \centering
      \includegraphics[width=\linewidth]{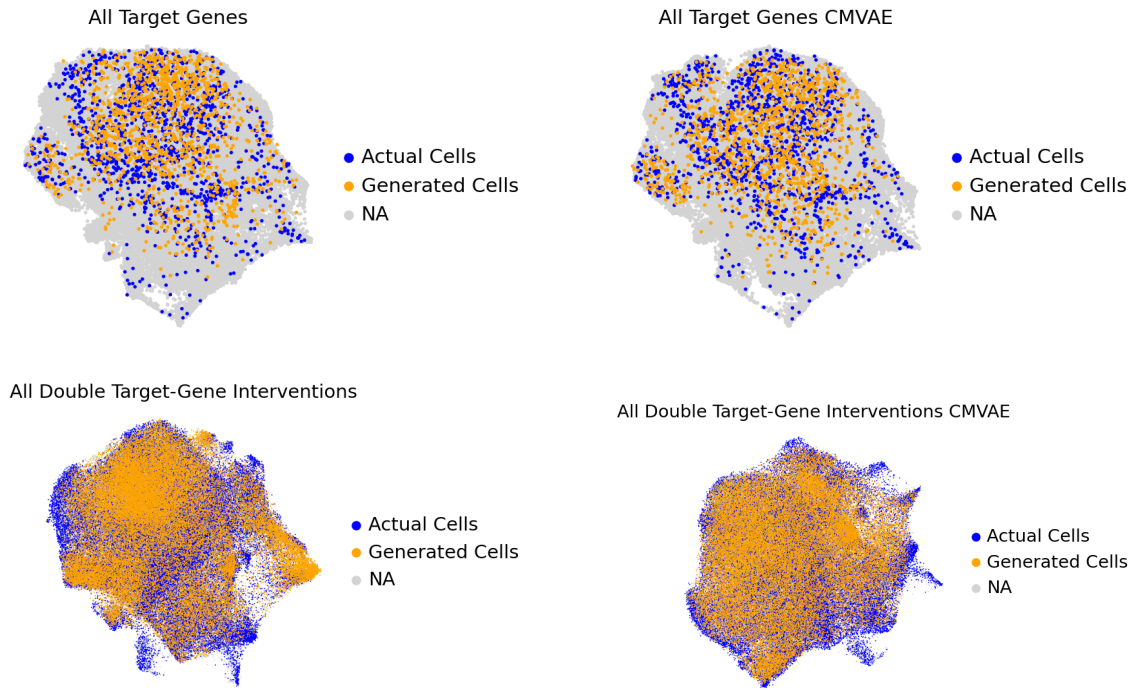}
    \end{minipage}
    \ifodd\i
      \hspace{0.04\textwidth}
    \else
      \par\vspace{3mm}
    \fi
  }
  \caption{Comparison of generated sample distributions under all single and all double interventions. We visualize the generated samples by \name{} (left) and CMVAE (right), alongside the actual perturbed cell distributions. }
  \label{fig:norman}
\end{figure*}

\begin{figure*}[htbp]
  \centering
  \foreach \i in {1,...,16} {
    \begin{minipage}[b]{0.23\textwidth}
      \centering
      \includegraphics[width=\linewidth]{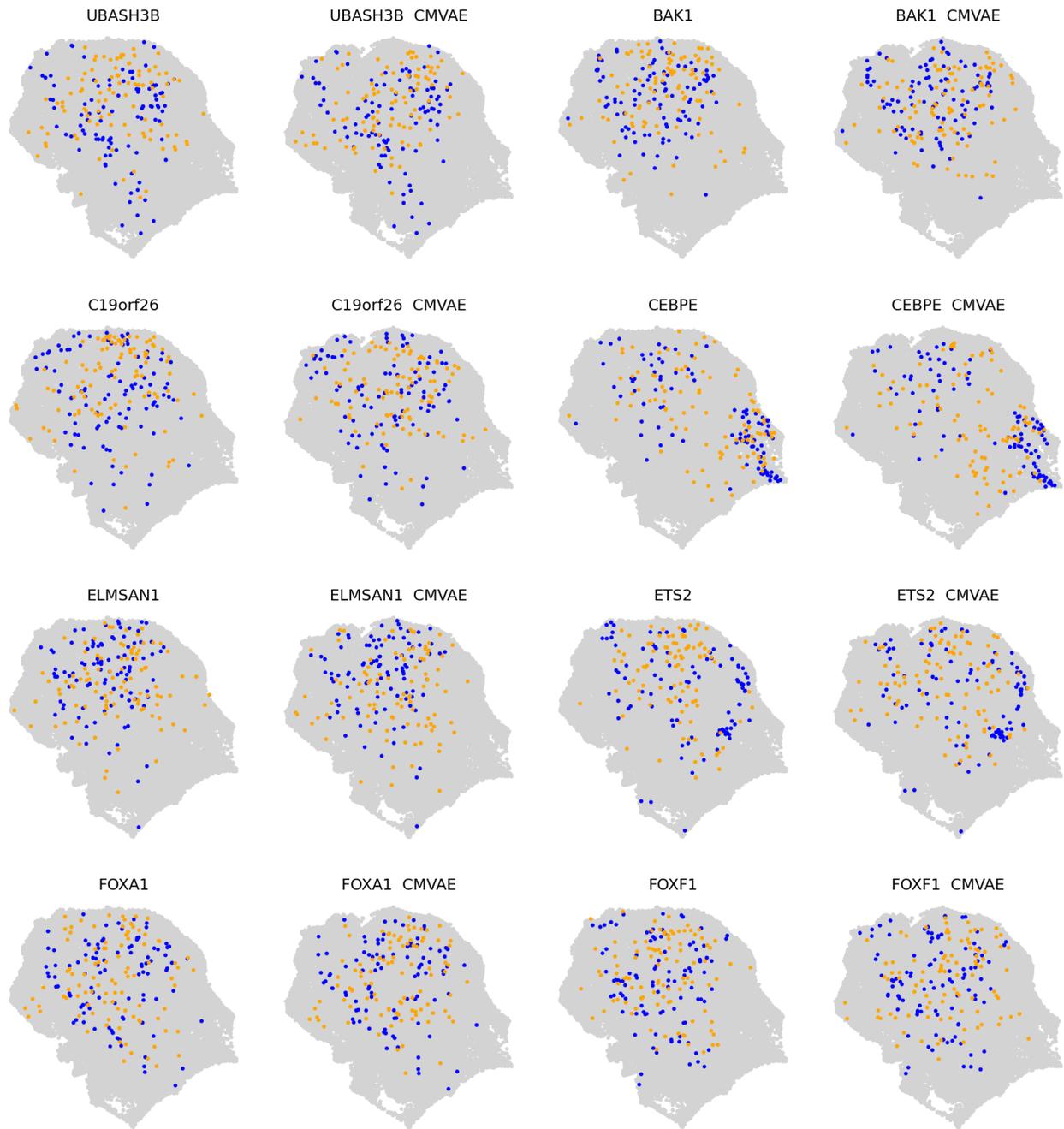}
    \end{minipage}
    \ifnum\i=28\relax\else
      \ifnum\i=0\relax\else
        \ifnum\i=4\relax\par\vspace{2mm}\fi
        \ifnum\i=8\relax\par\vspace{2mm}\fi
        \ifnum\i=12\relax\par\vspace{2mm}\fi
        \ifnum\i=16\relax\par\vspace{2mm}\fi
        \ifnum\i=20\relax\par\vspace{2mm}\fi
        \ifnum\i=24\relax\par\vspace{2mm}\fi
      \fi
    \fi
  }
  \caption{Comparison of generated samples (yellow) versus actual (or ground-truth) observed samples (blue) under single perturbations.  Each row corresponds to two pairs; each pair consists of \name{} (left) and CMVAE (right) alongside the actual perturbed cell distributions.}
  \label{fig:single1}
\end{figure*}

\begin{figure*}[htbp]
  \centering
  \foreach \i in {17,...,28} {
    \begin{minipage}[b]{0.23\textwidth}
      \centering
      \includegraphics[width=\linewidth]{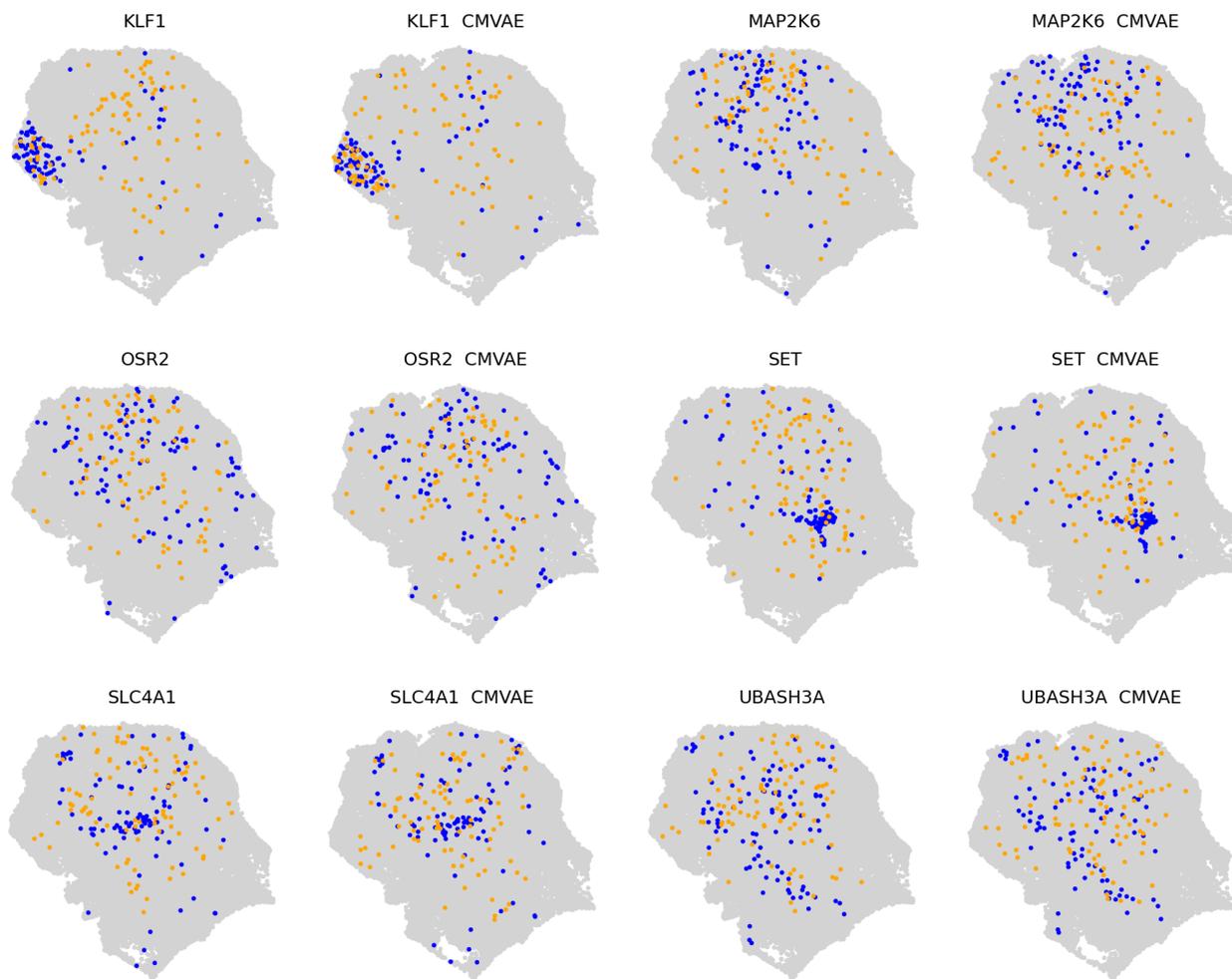}
    \end{minipage}
    \ifnum\i=28\relax\else
      \ifnum\i=0\relax\else
        \ifnum\i=4\relax\par\vspace{2mm}\fi
        \ifnum\i=8\relax\par\vspace{2mm}\fi
        \ifnum\i=12\relax\par\vspace{2mm}\fi
        \ifnum\i=16\relax\par\vspace{2mm}\fi
        \ifnum\i=20\relax\par\vspace{2mm}\fi
        \ifnum\i=24\relax\par\vspace{2mm}\fi
      \fi
    \fi
  }
  \caption{Comparison of generated samples (yellow) versus actual (or ground-truth) observed samples (blue) under single perturbations.  Each row corresponds to two pairs; each pair consists of \name{} (left) and CMVAE (right) alongside the actual perturbed cell distributions.}
  \label{fig:single2}
\end{figure*}

Figure~\ref{fig:norman} presents a comprehensive comparison between \name{} and CMVAE, showing both generated and real samples for all single and double interventions. Both models can broadly capture the effects of interventions. Figures~\ref{fig:single1} and~\ref{fig:single2} provide a detailed comparison for each of the 14 single interventions in the \textsc{norman} test set, illustrating close alignment between the generated and true sample distributions for both models. Notably, for certain interventions such as CEBPE and BAK1, \name{} achieves a noticeably closer match to the experimental data. Overall, by leveraging integrated network knowledge, \name{} consistently outperforms CMVAE, achieving higher $R^2$ scores and lower MMD losses across interventions.

\begin{figure*}[htbp]
  \centering

  \begin{minipage}[b]{0.23\textwidth}
    \centering
    \includegraphics[width=\linewidth]{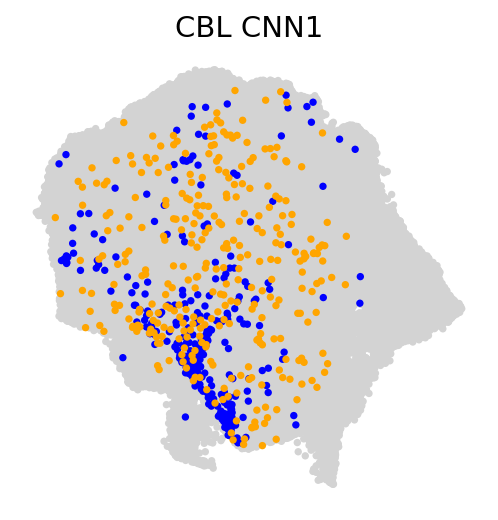}
  \end{minipage}\hfill
  \begin{minipage}[b]{0.23\textwidth}
    \centering
    \includegraphics[width=\linewidth]{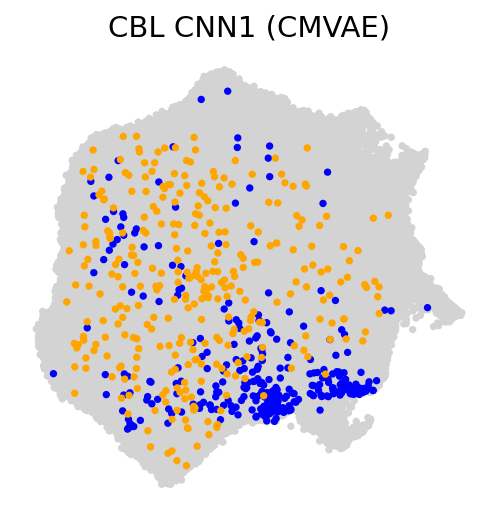}
  \end{minipage}\hfill
  \begin{minipage}[b]{0.23\textwidth}
    \centering
    \includegraphics[width=\linewidth]{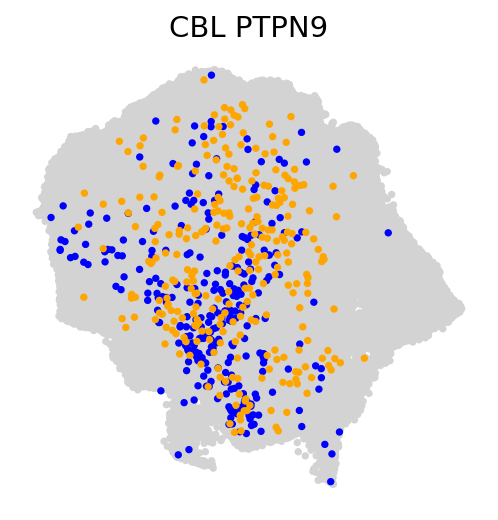}
  \end{minipage}\hfill
  \begin{minipage}[b]{0.23\textwidth}
    \centering
    \includegraphics[width=\linewidth]{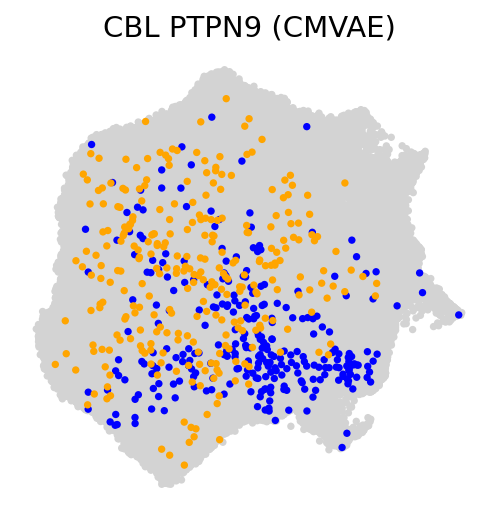}
  \end{minipage}

  \begin{minipage}[b]{0.23\textwidth}
    \centering
    \includegraphics[width=\linewidth]{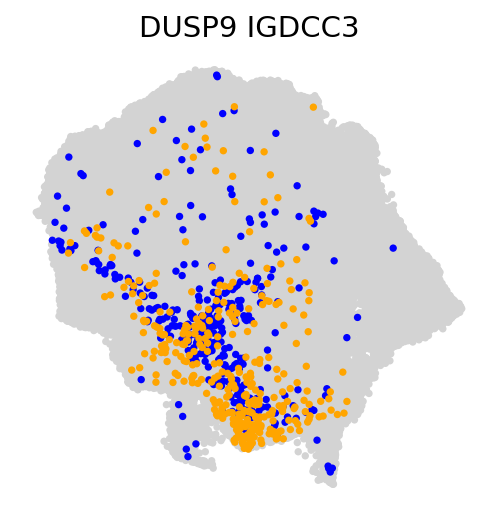}
  \end{minipage}\hfill
  \begin{minipage}[b]{0.23\textwidth}
    \centering
    \includegraphics[width=\linewidth]{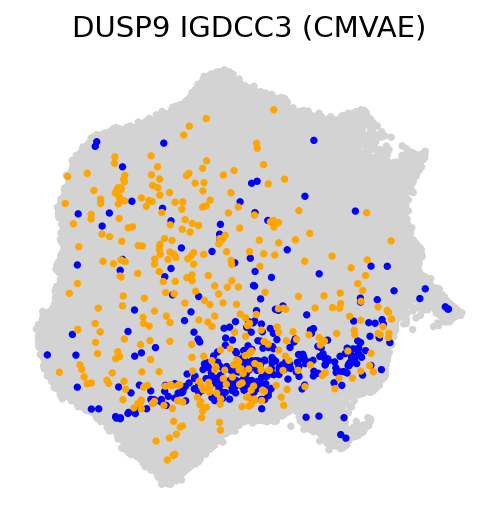}
  \end{minipage}\hfill
  \begin{minipage}[b]{0.23\textwidth}
    \centering
    \includegraphics[width=\linewidth]{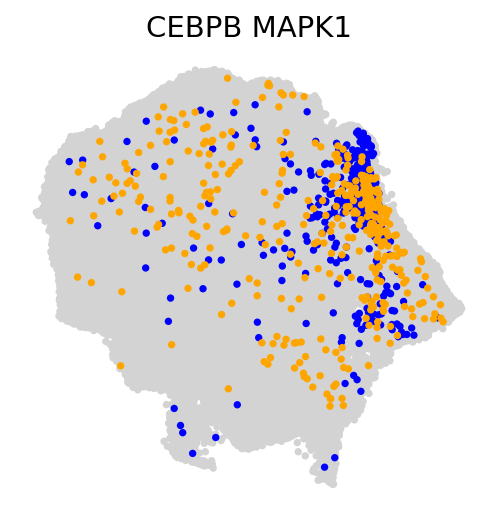}
  \end{minipage}\hfill
  \begin{minipage}[b]{0.23\textwidth}
    \centering
    \includegraphics[width=\linewidth]{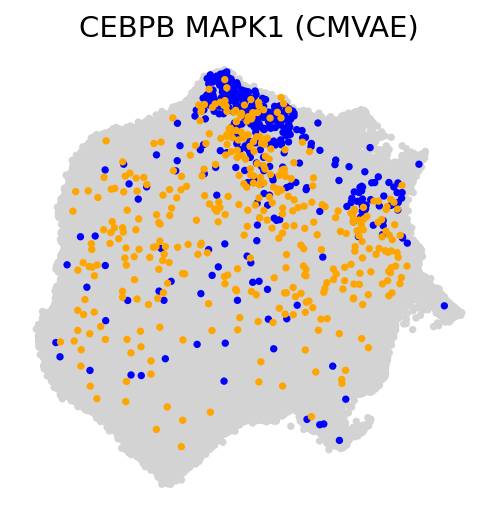}
  \end{minipage}

    \vspace{3mm}

  \begin{minipage}[b]{0.23\textwidth}
    \centering
    \includegraphics[width=\linewidth]{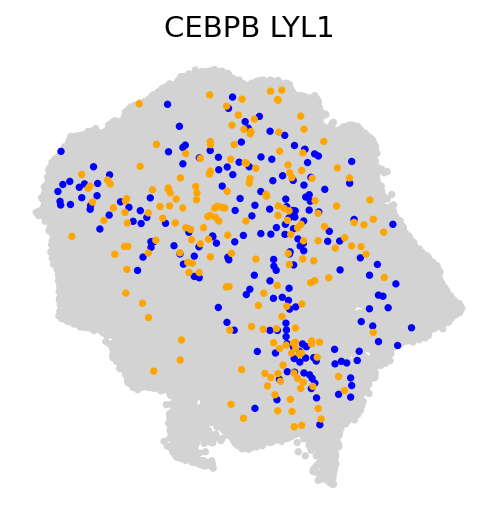}
  \end{minipage}\hfill
  \begin{minipage}[b]{0.23\textwidth}
    \centering
    \includegraphics[width=\linewidth]{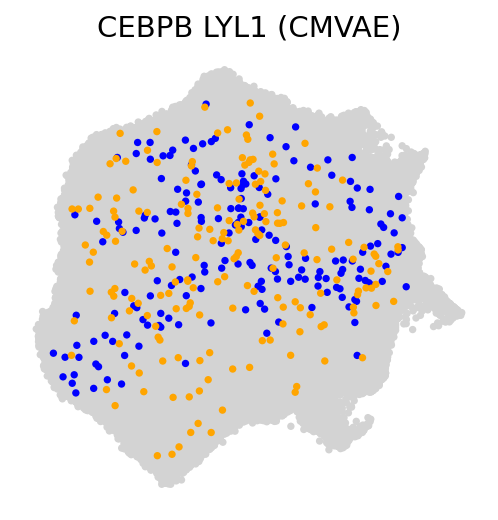}
  \end{minipage}\hfill
  \begin{minipage}[b]{0.23\textwidth}
    \centering
    \includegraphics[width=\linewidth]{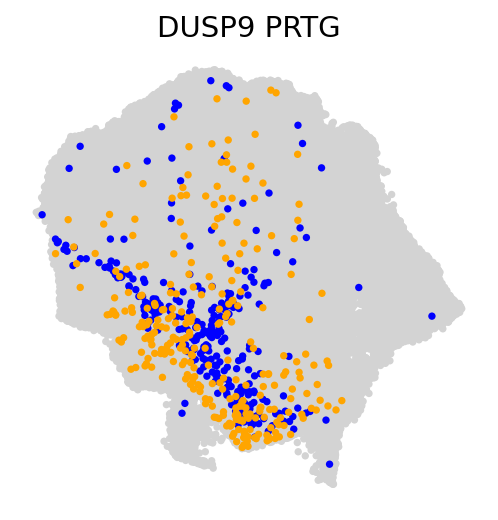}
  \end{minipage}\hfill
  \begin{minipage}[b]{0.23\textwidth}
    \centering
    \includegraphics[width=\linewidth]{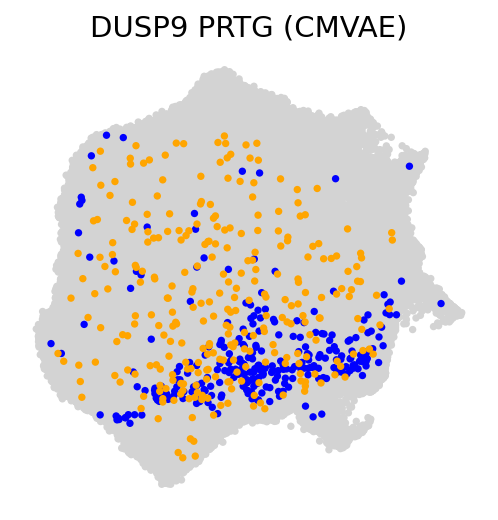}
  \end{minipage}
  
  \begin{minipage}[b]{0.23\textwidth}
    \centering
    \includegraphics[width=\linewidth]{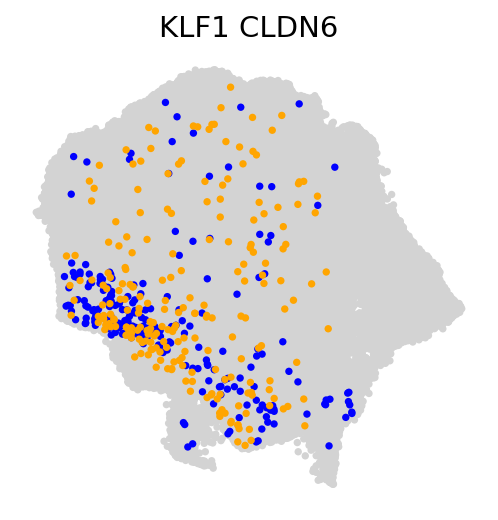}
  \end{minipage}\hfill
  \begin{minipage}[b]{0.23\textwidth}
    \centering
    \includegraphics[width=\linewidth]{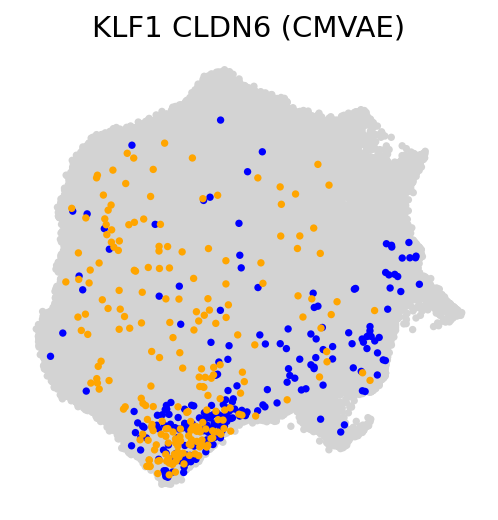}
  \end{minipage}\hfill
  \begin{minipage}[b]{0.23\textwidth}
    \centering
    \includegraphics[width=\linewidth]{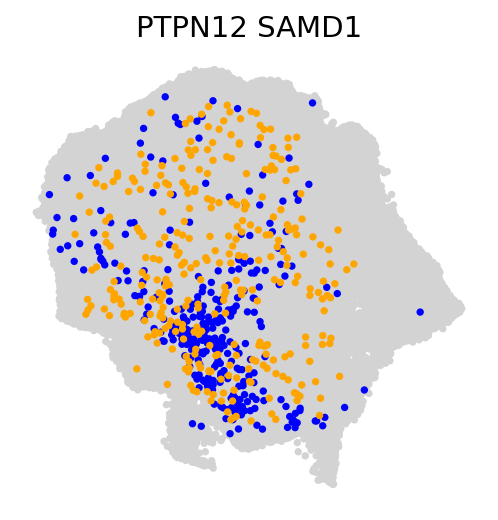}
  \end{minipage}\hfill
  \begin{minipage}[b]{0.23\textwidth}
    \centering
    \includegraphics[width=\linewidth]{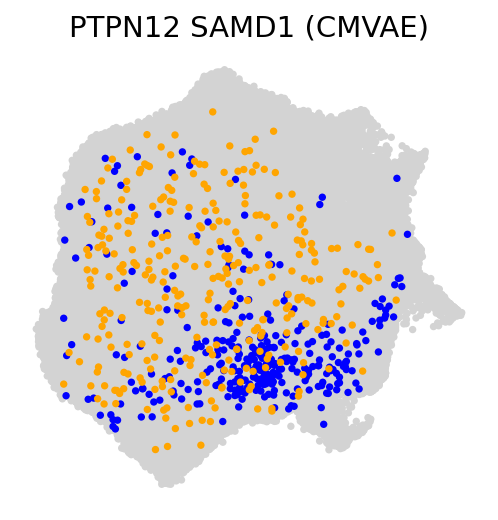}
  \end{minipage}

  \caption{Comparison of generated samples (yellow) versus actual (or ground-truth) observed samples (blue) under double interventions.  Each row corresponds to two pairs; each pair consists of \name{} (left) and CMVAE (right) alongside the actual perturbed cell distributions.}
  \label{fig:double1}
\end{figure*}

\begin{figure*}[htbp]
  \centering

  \begin{minipage}[b]{0.23\textwidth}
    \centering
    \includegraphics[width=\linewidth]{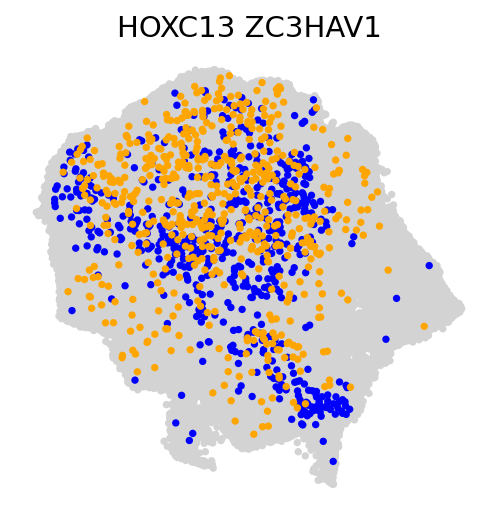}
  \end{minipage}\hfill
  \begin{minipage}[b]{0.23\textwidth}
    \centering
    \includegraphics[width=\linewidth]{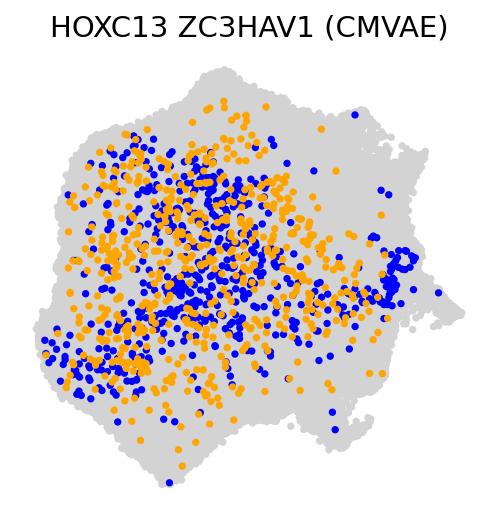}
  \end{minipage}\hfill
  \begin{minipage}[b]{0.23\textwidth}
    \centering
    \includegraphics[width=\linewidth]{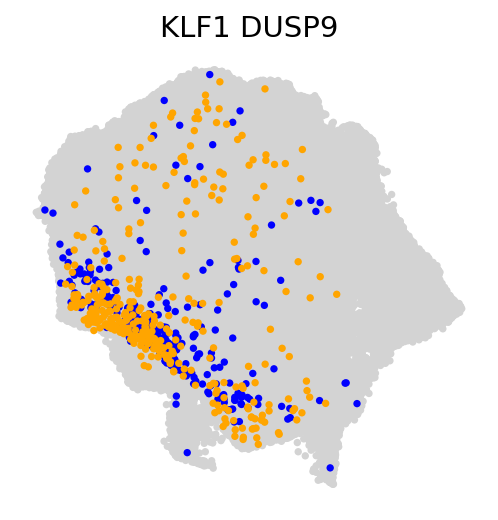}
  \end{minipage}\hfill
  \begin{minipage}[b]{0.23\textwidth}
    \centering
    \includegraphics[width=\linewidth]{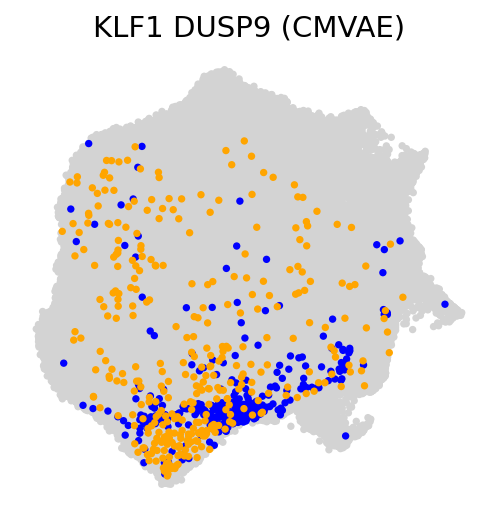}
  \end{minipage}

  \vspace{3mm}

  \begin{minipage}[b]{0.23\textwidth}
    \centering
    \includegraphics[width=\linewidth]{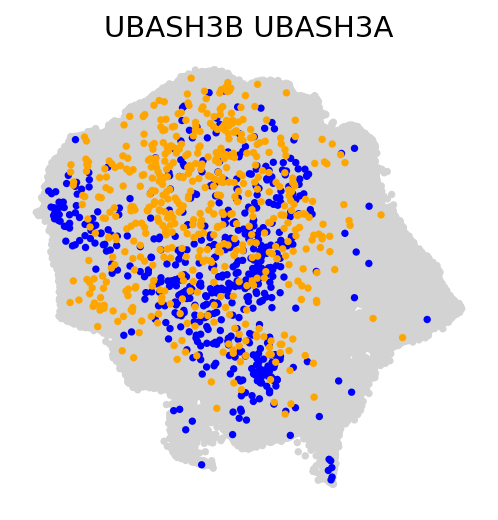}
  \end{minipage}\hfill
  \begin{minipage}[b]{0.23\textwidth}
    \centering
    \includegraphics[width=\linewidth]{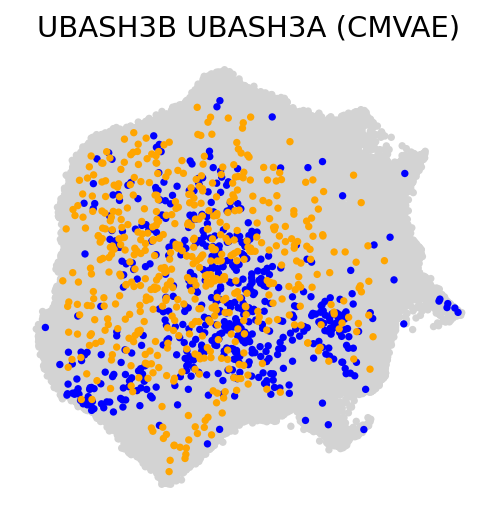}
  \end{minipage}\hfill
  \begin{minipage}[b]{0.23\textwidth}
    \centering
    \includegraphics[width=\linewidth]{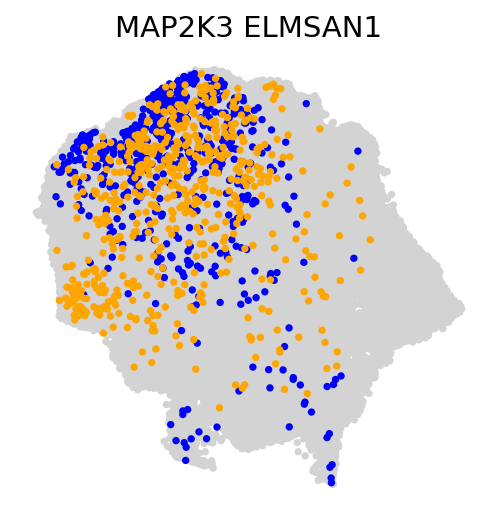}
  \end{minipage}\hfill
  \begin{minipage}[b]{0.23\textwidth}
    \centering
    \includegraphics[width=\linewidth]{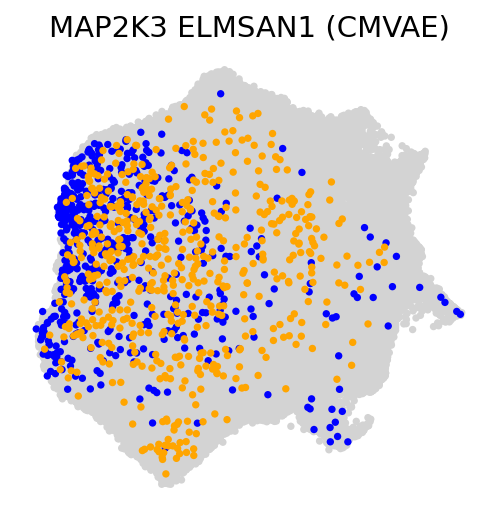}
  \end{minipage}

    \vspace{3mm}

  \begin{minipage}[b]{0.23\textwidth}
    \centering
    \includegraphics[width=\linewidth]{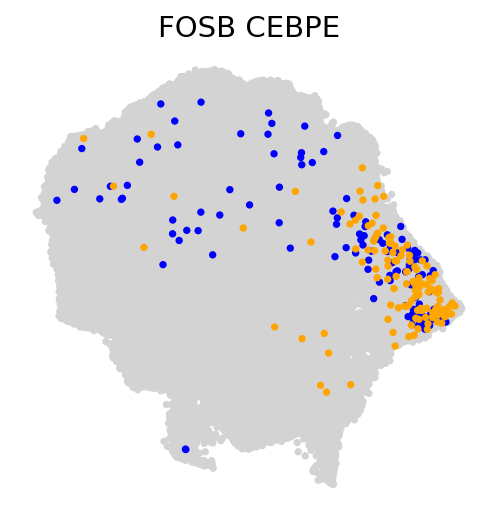}
  \end{minipage}\hfill
  \begin{minipage}[b]{0.23\textwidth}
    \centering
    \includegraphics[width=\linewidth]{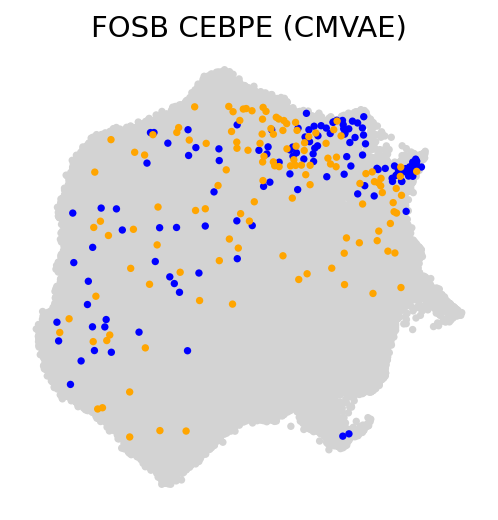}
  \end{minipage}\hfill
  \begin{minipage}[b]{0.23\textwidth}
    \centering
    \includegraphics[width=\linewidth]{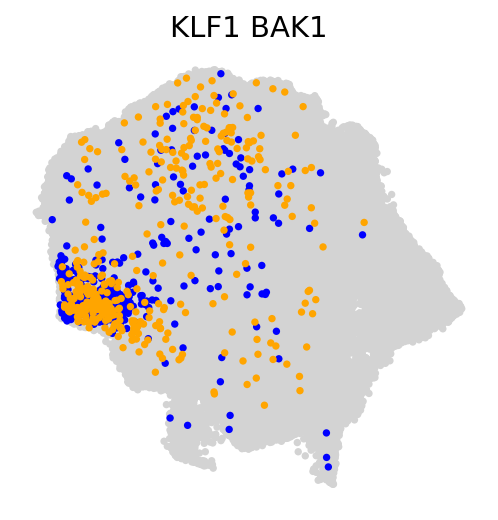}
  \end{minipage}\hfill
  \begin{minipage}[b]{0.23\textwidth}
    \centering
    \includegraphics[width=\linewidth]{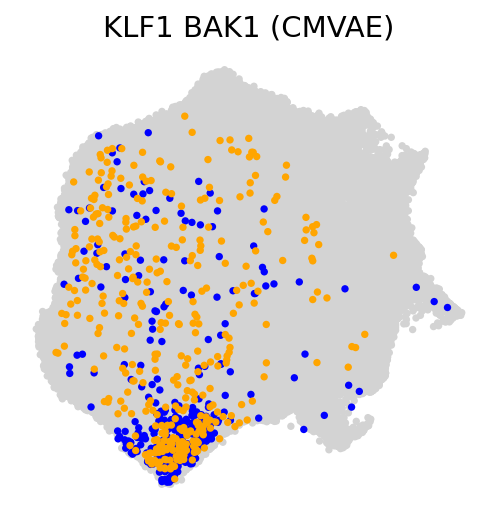}
  \end{minipage}
  
  \begin{minipage}[b]{0.23\textwidth}
    \centering
    \includegraphics[width=\linewidth]{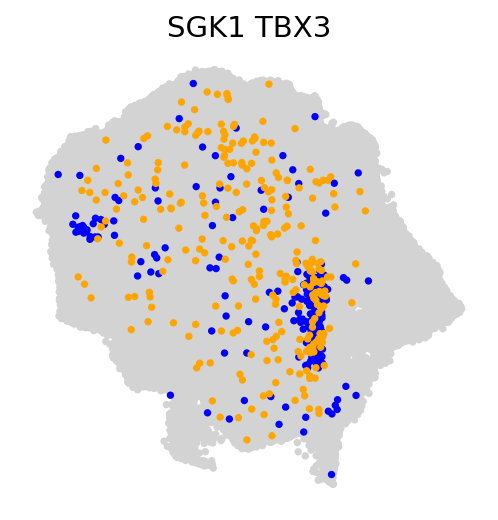}
  \end{minipage}\hfill
  \begin{minipage}[b]{0.23\textwidth}
    \centering
    \includegraphics[width=\linewidth]{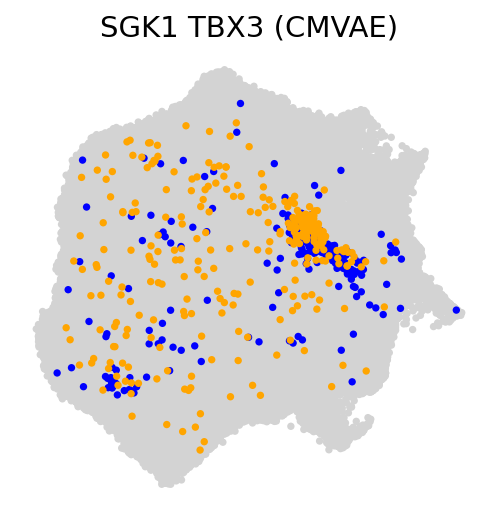}
  \end{minipage}\hfill
  \begin{minipage}[b]{0.23\textwidth}
    \centering
    \includegraphics[width=\linewidth]{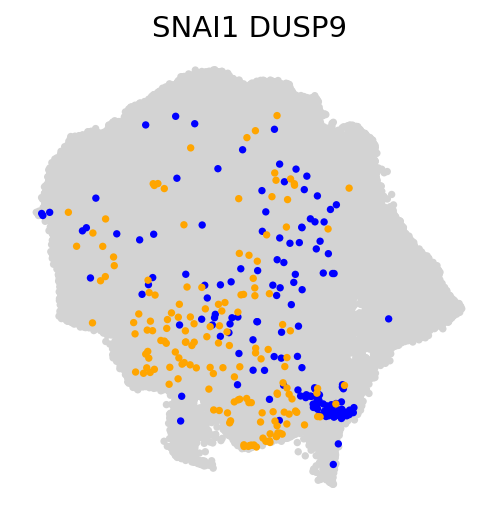}
  \end{minipage}\hfill
  \begin{minipage}[b]{0.23\textwidth}
    \centering
    \includegraphics[width=\linewidth]{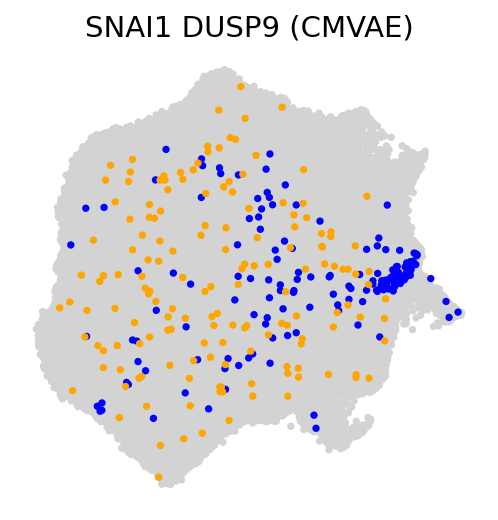}
  \end{minipage}

  \caption{Comparison of generated samples (yellow) versus actual (or ground-truth) observed samples (blue) under double interventions.  Each row corresponds to two pairs; each pair consists of \name{} (left) and CMVAE (right) alongside the actual perturbed cell distributions.}
  \label{fig:double2}
\end{figure*}

Figures~\ref{fig:double1} and~\ref{fig:double2} present representative results for double intervention distributions. In many cases, \name{} more accurately captures the effects of double perturbations compared to CMVAE. However, there are certain double interventions where neither \name{} nor CMVAE fully recapitulates the true data distribution, highlighting ongoing challenges in modeling complex combinatorial effects.

\begin{figure*}[t]
\centering
\includegraphics[width=\linewidth]{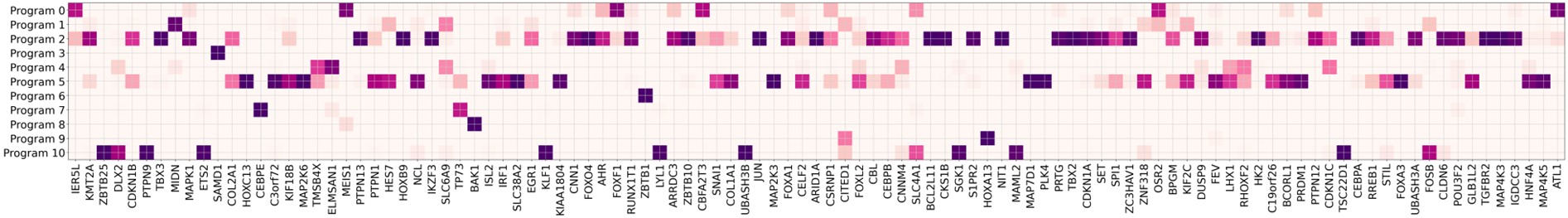}
\caption{A detailed visualization of the latent program for \textsc{norman}, illustrating for each latent program the full set of associated target gene.}
\label{fig:program}
\end{figure*}

\end{document}